\documentclass[letterpaper, 10 pt, conference, table]{ieeeconf}  %

\IEEEoverridecommandlockouts                              %
\makeatletter
\let\NAT@parse\undefined
\makeatother

\PassOptionsToPackage{usenames, dvipsnames}{xcolor}
\usepackage[usenames,dvipsnames,rgb]{xcolor}

\newcommand*\linkcolours{ForestGreen}

\usepackage{times}
\usepackage{graphicx}
\usepackage{amssymb}
\usepackage{gensymb}
\usepackage{amsmath}
\usepackage{breakurl}

\usepackage{url,hyperref}
\usepackage[hypcap=true]{caption}
\usepackage{booktabs, makecell}

\usepackage{threeparttable}
\hypersetup{
colorlinks,
linkcolor=\linkcolours,
citecolor=\linkcolours,
filecolor=\linkcolours,
urlcolor=\linkcolours}

\usepackage{algpseudocode}
\usepackage{algorithm}

\algnewcommand\algorithmicforeach{\textbf{for each}}
\algdef{S}[FOR]{ForEach}[1]{\algorithmicforeach\ #1\ \algorithmicdo}

\usepackage[labelfont={bf},font=small]{caption}
\usepackage[none]{hyphenat}

\usepackage{mathtools, cuted}

\usepackage[noadjust, nobreak]{cite}

\usepackage{tabularx}
\usepackage{amsmath}

\usepackage{float}

\usepackage{pifont}%

\newcolumntype{Y}{>{\centering\arraybackslash}X}

\usepackage[]{placeins}

\usepackage{placeins}

\usepackage{tikz}

\usepackage[framemethod=tikz]{mdframed}

\usepackage{afterpage}

\usepackage{stfloats}

\usepackage{rotating}
\usepackage{atbegshi}
\newcommand{\handlethispage}{}
\newcommand{\discardpagesfromhere}{\let\handlethispage\AtBeginShipoutDiscard}
\newcommand{\keeppagesfromhere}{\let\handlethispage\relax}
\AtBeginShipout{\handlethispage}

\usepackage{comment}
\usepackage{bm}
\renewcommand{\vec}[1]{\boldsymbol{\mathbf{#1}}}

\def\Ac{\mathcal{A}}
\def\Sc{\mathcal{S}}

\def\Mc{\mathcal{M}}

\def\Rc{\mathcal{R}}
\definecolor{emerald}{rgb}{0.31, 0.78, 0.47}

\usepackage[T1]{fontenc}
\usepackage{inputenc}
\usepackage[english]{babel}
\usepackage{collcell}
\usepackage{hhline}
\usepackage{pgf}
\usepackage{multirow}
\usepackage[numbers]{natbib}

\def\colorModel{hsb} %
\newcommand\ColCell[1]{
	\pgfmathparse{#1<50?1:0}  %
	\ifnum\pgfmathresult=0\relax\color{white}\fi
	\pgfmathsetmacro\compA{0}      %
	\pgfmathsetmacro\compB{#1/100} %
	\pgfmathsetmacro\compC{1}      %
	\edef\x{\noexpand\centering\noexpand\cellcolor[\colorModel]{\compA,\compB,\compC}}\x #1
}
\newcolumntype{E}{>{\collectcell\ColCell}m{0.4cm}<{\endcollectcell}}  %

\DeclareMathOperator*{\argmax}{argmax}

\title{\LARGE \bf
Learning to Solve Vehicle Routing Problems: A Survey
}

\author{ Aigerim Bogyrbayeva$^\dag$,
Meraryslan Meraliyev $^\flat$, Taukekhan Mustakhov$^\dag$, Bissenbay Dauletbayev$^\dag$
\thanks{$^\dag$Department of Computer Science, Suleyman Demirel University (SDU), Qaskelen, Kazakhstan Email: {\tt\footnotesize\{aigerim.bogyrbayeva,  taukekhan.mustakhov, b.dauletbayev\}@sdu.edu.kz}}
\thanks{$^\flat$ corresponding author. Department of Computer Science, Suleyman Demirel University (SDU), Qaskelen, Kazakhstan Email: {\tt\footnotesize meraryslan.meraliyev@sdu.edu.kz}}
\thanks{This work is supported by the Computer Science Department funding and the internal grant of SDU for 2022-2023.}
}

\begin{document}

\maketitle
\thispagestyle{plain}
\pagestyle{plain}

\begin{abstract}
This paper provides a systematic overview of machine learning methods applied to solve NP-hard Vehicle Routing Problems (VRPs). Recently, there has been a great interest from both machine learning and operations research communities to solve VRPs either by pure learning methods or by combining them with the traditional hand-crafted heuristics. We present the taxonomy of the studies for learning paradigms, solution structures, underlying models, and algorithms. We present in detail the results of the state-of-the-art methods demonstrating their competitiveness with the traditional methods. The paper outlines the future research directions to incorporate learning-based solutions to overcome the challenges of modern transportation systems. 
\end{abstract}


\section{Introduction}
Cost-effective logistics systems overall define the competitiveness of the companies, and the relation of logistics expenditure to GDP indicates the effectiveness of business operations in a country. For instance, in 2018, the USA businesses spent 10.4\% of their revenue on transportation costs alone, while the overall logistics expenditure constituted 8\% of GDP \cite{logreport}. 
Along with increased fuel prices, the transportation costs are mainly influenced by the last-mile delivery, which is defined as transporting goods from a warehouse to a customer's location. With the increased demand for online sales, the last mile delivery system's effectiveness has become essential as it substitutes for 50\% of the total transportation costs \cite{yuan2018last}. Also, the carbon dioxide footprint is another emerging concern in logistics systems. In order to overcome the above-outlined challenges, efficiently solving vehicle routing problems has been of great interest to both practitioners and researchers. 

The Vehicle Routing Problem, in general, are defined in either a directed or undirected graph $\mathbb{G}(\mathbb{V}, \mathbb{E})$ consisting of a set of nodes $\mathbb{V} = \{v_0, \dots, v_n \}$ and a set of edges or arcs. We focus on a undirected graph with a set of edges, $\mathbb{E} = \{(v_i, v_j): i < j, v_i, v_j \in \mathbb{V} \}$. VRP aims to construct routes with the minimal total cost for $M$ number of identical vehicles leaving from a depot, $v_0$, to visit all the nodes,$\mathbb{V} \setminus v_0$, representing customers with non-negative demand $d_v$, and return to the depot, where each customer is visited only once. Different constraints are added to VRP to reflect the realistic routing challenges faced by delivery companies \cite{rizzoli2007ant}. In Capacitated Vehicle Routing Problem (CVRP), each vehicle is subjected to maximum capacity $Q_m$ such that the total demand of visited customers in its route does not exceed $Q_m$. In Vehicle Routing Problems with Time Windows (VRPTW), vehicles must arrive at customer location within specified time windows. However, all the above variations of the VRP belong to vertex routing problems. Some applications, such as mail delivery and bus routing, are concerned with visiting arcs called arc routing problems (ARPs). 

As generalizations of Travelling Salesman Problem (TSP), VRPs belong to the family of combinatorial optimization problems and are known to be NP-hard  \cite{lenstra1981complexity}. Therefore, approximation or heuristic algorithms have been developed to solve large size VRPs \cite{laporte1992vehicle, doerner2010survey, asghari2021green}, while Mixed Integer Programming (MIP) is used to find exact solutions on small instances. The hand-crafted heuristics are built on the experts' domain knowledge about the specifics of the structure of the problem at hand. Even though such heuristics often produce solutions in a short amount of time, they fail to generalize to slight changes in the inputs and must be solved from scratch \cite{bogyrbayeva2021reinforcement}. 

\begin{figure}
    \centering
    \includegraphics[scale=0.6]{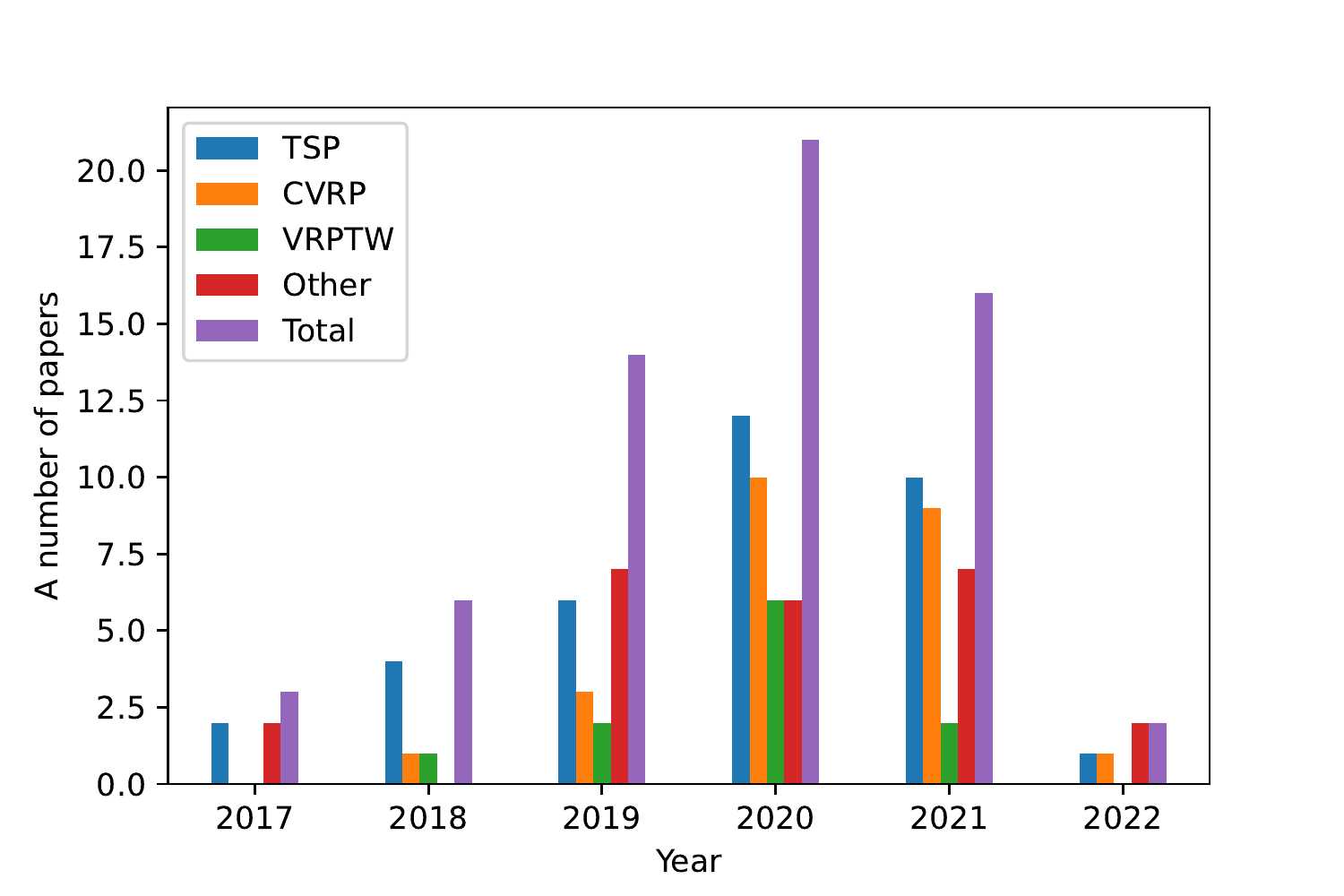}
    \caption{A number of studies focused on solving VRPs using learning methods.}
    \label{fig:papers}
\end{figure}

On the other hand, machine learning methods have shown great success across many applications due to advancements in algorithms and hardware. Machine Learning, in general, can be viewed as applied statistics, where the computational power of computers is used to \emph{learn} approximately functions instead of explicitly writing computer programs. The main advantage of learning models is their ability to generalize to input problems coming from identical distributions and produce solutions instantaneously \cite{nazari2018reinforcement}. Therefore, there has been a surge of studies in recent years aiming to develop novel models and training algorithms for routing problems using machine-learning methods to achieve near-optimal solutions. The figure \ref{fig:papers} indicates the increasing trend in the number of such studies.

There are two main learning paradigms used for solving VRPs. In Supervised Learning, a model is trained on a training dataset consisting of input problems and corresponding solutions. In Reinforcement Learning, VRPs are sequential decision-making problems and rely on the Markov Decision Process (MDP) to construct solutions. However, pure learning-based heuristics have disadvantages such as generalization, poor solution quality, and scalability, which led to new methods that combine the traditional heuristics with learning methods. An increasingly growing literature has attracted the attention of both machine learning and operations research communities \cite{bengio2021machine}.

Therefore, this survey aims to systematically present the recent advancements in learning methods to solve routing problems and discuss the challenges and opportunities for future research. We believe the survey will benefit researchers interested in solving VRPs either using pure learning methods or hybrid methods that combine learning and construction heuristics.

Compared to the existing surveys \cite{ritzinger2016survey, cappart2021combinatorial, mazyavkina2021reinforcement, vesselinova2020learning}, the presented study does not limit its scope to Reinforcement Learning methods only but surveys all studies that used any learning paradigm. Also, we focus our attention on routing problems in particular rather than discussing the combinatorial optimization problems in general. We consider our work to be complementary to the existing studies.

Our contributions can be summarized as follows:
\begin{itemize}
    \item This is the first survey focused on learning methods to solve routing problems in general.
    \item We present taxonomy to classify the main research directions in the existing literature that includes both Supervised and Reinforcement Learning Methods.
    \item We provide detailed overview of the hybrid models that combine the hand-crafted heuristics with learning methods. 
\end{itemize}

We have the following outline for the paper: in Section \ref{sec:background}, we provide an overview of learning paradigms, models, and algorithms; in Section \ref{sec:taxonomy}, we present the central taxonomy for learning methods to solve routing problems and discuss end-to-end learning methods; in Section \ref{sec:learnhybrid}, we focus on hybrid methods and present different solution structures; in Section \ref{sec:forms} we present single and multiple VRP formulations and results; in Section \ref{sec:future}, we discuss the future research directions and finalize with the concluding remarks.

\section{Background}\label{sec:background}
This section discusses two learning paradigms applied to solve VRPs, namely Supervised Learning and Reinforcement Learning. Unlike the former, Reinforcement Learning does not rely on given labels but instead focuses on learning through interactions in order to maximize a long-term reward. A  goal-oriented reinforcement learning \emph{agent} collects experiences by interacting with \emph{environment}, where the interaction dynamics describe a task at hand. The agent aims to learn decisions that will lead to the most significant total reward through trial and error. We introduce Markov Decision Process and reinforcement learning algorithms that solve sequential decision-making problems. 

\subsection{Supervised Learning}
In supervised learning, we are given a vector of inputs, $X^T = (X_1,X_2,\dots, X_p)$ consisting of $p$ number of features and a vector of labels, $Y$. Then we can assemble a training set $D = (x_i, y_i), i=1, \dots, N$ and  $x_i \in \mathbb{R}^p$. A supervised learning algorithm takes input $x_i$ and produces the predicted values of output denoted as $\hat{f}(x_i)$. The algorithm is trained by adjusting its predictions to reduce the difference between the actual output values, $y_i$, and its predicted values, $\hat{f}(x_i)$. Depending on the problem, we can define different loss functions to turn training parameters $\theta$ efficiently. For the recent survey of the loss function, please see \cite{wang2022comprehensive}.
After training, the learned function $\hat{f}(x)$ is used to predict the values of output with new values of $x$.

\subsection{Markov Decision Process (MDP)}
Formally, reinforcement learning can be described in detail with MDP. 
Finite MDP consists of a tuple $\Mc ={\langle}\Sc, \Ac, \Rc, P {\rangle}$ representing a finite set of states, actions,  rewards and a transition probability function, respectively. At time $t$, given the current state of environment $S_t$,  an agent selects action $A_t$ that yields reward in the next step denoted by $R_{t+1}$ and a new state $S_{t+1}$. The state-transition probability function given in \ref{eq:state_p} defines the dynamics of a problem, where the next state of the environment is only determined by the current state and action showcasing the Markov property: 
\begin{align}\label{eq:state_p}
    p(s'|s, a) = P(S_{t+1}=s'|S_t=s, A_t=a). 
\end{align}
The reward function determines the expected reward for the taken action $a$ at state $s$ with the consecutive state $s'$:
\begin{align}
    r(s, a, s') = \mathbb{E}[R_{t+1}| S_t=s, A_t=a, S_{t+1}=s']. \label{eq:reward}
\end{align}
In MDPs, a policy denoted as $\pi(a|s)$ maps from states to probabilities of selected actions as shown below:
\begin{align}
    \pi(a|s) = p(A_{t+1} = a | S_t=s). \label{eq:trans_prob}
\end{align}
The value function, $v(s)$, relates states to long-term rewards. Formally a values function under policy $\pi$ denoted as $v_{\pi}(s)$ is the expected total reward starting from state $s$ and following policy $\pi$ with planning horizon $T$ and  discount factor $\gamma$:
\begin{align}
    v_{\pi}(s) = \mathbb{E}[\sum_{k=0}^{T-1}\gamma^kR_{t+k+1}|S_t=s] 
\end{align}
Similarly, action-value function $q_\pi(s, a)$ defines the expected reward when taking action $a$ at state $s$ under policy $\pi$:
\begin{align}
    q_{\pi}(s,a) = \mathbb{E}[\sum_{k=0}^{T-1}\gamma^kR_{t+k+1}|S_t=s, A_t=a.] 
\end{align}
We are often interested in finding a policy that leads to the most significant expected total rewards. In other words, we solve for an optimal policy $\pi^*$ that yields the optimal state-value function $v^*(s)$:
\begin{align}\label{eq:optv}
    v^*(s) = & \max_{\pi} v_\pi(s)= \\ \nonumber 
    & \max_a\mathbb{E}[R_{t+1}+\gamma v^*(S_{t+1})|S_t=s, A_t=a] 
\end{align}
or the optimal action-value function $q^*(s,a)$:
\begin{align}\label{eq:optq}
    q^*(s, a) = & \max_{\pi} q_\pi(s, a)= \\ \nonumber &\max_{a^{'}}\mathbb{E}[R_{t+1}+\gamma q^*(S_{t+1}, a')|S_t=s, A_t=a] 
\end{align}
Equations \eqref{eq:optv} - \eqref{eq:optq} are known as the Bellman optimality equations that are recursive.

\subsection{Reinforcement Learning Algorithms}
\begin{figure*}
    \centering
    \includegraphics[scale=0.5]{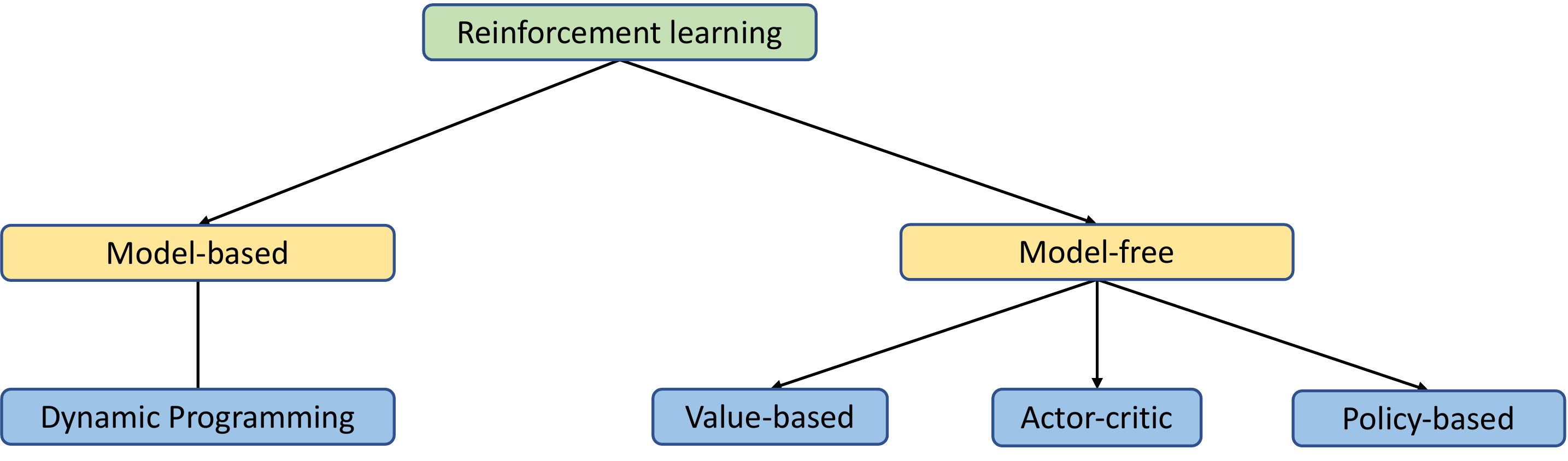}
    \caption{An overview of reinforcement learning methods.}
    \label{fig:rl}
\end{figure*}

We can broadly define \emph{reinforcement learning} as any method that solves the above-defined MDPs. 
When Equations \eqref{eq:reward} and \eqref{eq:trans_prob}  are known, one can directly compute the future reward and apply planning methods. The methods that require complete knowledge about the dynamics of the problems are called \emph{model-based} and dynamic programming is one of such methods. Often Equations \eqref{eq:reward} and \eqref{eq:trans_prob} are unknown, and one needs to learn the dynamics of the problem through experiences. The methods that focus directly on finding optimal state-value functions rather than requiring the complete knowledge of transition probabilities are called \emph{model-free}. Figure \ref{fig:rl} provides an overview of reinforcement learning. Further, model-free methods can be categorized into value-based, actor-critic, and policy-based, depending on which function they aim to optimize.

Typically, in value-based methods, we aim to find the action-value function, $q^*$. One of the widely used algorithms for such purpose is \emph{Q-learning} \cite{watkins1992q} that recursively updates the values of action and state pairs independently from a policy with $\alpha$ as a constant step size:
\begin{align}\label{eq:q}
    Q(S_t, A_t) \gets & Q(S_t, A_t) + \\ \nonumber
    &+ \alpha [R_{t+1}+\gamma \max_a Q(S_{t+1}, a) - Q(S_t, A_t)]
\end{align}
In the simplest form shown above, Q-learning takes as a target one step Temporal-Difference (TD), $R_{t+1}+\gamma \max_a Q(S_{t+1}, a)$. The algorithm uses policy $\pi$ to select actions and states whose values will be updated; however, the updates of Q-values are done through maximization over actions. To balance exploration and exploitation, Q-learning deploys $\epsilon$-greedy policy that selects random actions with probability $\epsilon$ and follows the greedy policy concerning the current estimates of Q with probability $1-\epsilon$. 

Methods that directly focus on finding optimal policies are called policy-based methods. 
For instance, in policy-gradient methods, we directly aim to learn a parametrized policy $\pi_\theta$ with a set of parameters $\theta$ defined as follows:
\begin{align}
    \pi_\theta(a|s) = Pr(A_t=a | S_t=s, \theta=\theta_t) 
\end{align}
To find the optimal values of parameters $\theta$, we need to define an objective function $J(\theta)$, which we aim to maximize. Then we need to solve the optimization problem with the gradient ascent to update the values of $\theta$ with step size $\alpha$:
\begin{align}
    \theta \gets \theta + \alpha \nabla_{\theta}J(\theta)
\end{align}
For instance, the objective function can be set to maximize the total reward at the end of the episode, $R_T$ then according to the policy gradient theorem \cite{sutton1999policy}  we have:
\begin{align}\label{eq:reinforce}
    \nabla_{\theta}J(\theta) = \mathbb{E}[\nabla_{\theta}\text{log}\pi_{\theta}(a|s)R_T]
\end{align}
In fact, the REINFORCE algorithm \cite{williams1992simple} widely adopted to solve VRPs  uses Monte-Carlo gradient methods as in Equation \eqref{eq:reinforce} to learn optimal policies. However, using the total reward of an episode results in high variance, which can be mitigated by the advantage function defined as the difference between the total reward and baseline:
\begin{align}\label{eq:reinforce}
    \nabla_{\theta}J(\theta) = \mathbb{E}[\nabla_{\theta}\text{log}\pi_{\theta}(a|s)(R_T-b)]
\end{align}
In practice,  actor-critic methods are often used, where an actor produces policy, and a critic is responsible for the policy evaluation. In contrast to Monte-Carlo gradient methods, TD is used to update values in actor-critic policy gradients.

In recent years neural networks that, according to the universal approximation theorem, can approximate any continuous function with mild assumptions \cite{cybenko1989approximation, hornik1991approximation} have been combined with the reinforcement learning algorithms mentioned above that gave rise to  \emph{Deep Reinforcement Learning} (DRL). For instance, in DRL, two separate neural networks are trained to estimate policy and state-value functions according to actor-critic methods \cite{mnih2016asynchronous,lillicrap2015continuous, schulman2017proximal}. Similarly, new algorithms have been proposed to use neural networks for Q-learning with high-dimensional data \cite{mnih2015human, van2016deep, schaul2015prioritized}.

\subsection{Multi-agent Reinforcement Learning}
The algorithms, as mentioned earlier, focus on finding optimal policies for a single agent. However, real-life applications such as VRPs require learning policies for a fleet of vehicles. In general, sequential decision problems that involve several agents that either are fully competing, fully collaborating, or with mixed strategies can be formulated with Multi-agent Reinforcement Learning (MARL). Unlike single-agent reinforcement learning, in MARL, due to the presence of several agents in a shared environment, the stationary assumption for MDPs is violated \cite{de2017bounding}. Therefore, recently developed deep MARL algorithms \cite{lowe2017multi, foerster2018counterfactual, omidshafiei2017deep} use Partially Observable MDP, which allows for each agent to have a local observation of the environment. Another challenge of MARL is scalability coming from increased complexity with the number of agents \cite{hernandez2019survey}. Along with the theoretical challenges of MARL, communication between agents becomes essential in the setting of many applications that may require training additional models to pass messages between agents \cite{zhang2021multi}.

\section{Taxonomy of Learning Methods for Routing Problems} \label{sec:taxonomy}
No single classification will cover the wide range of learning methods applied to solve routing problems. However, one broad classification can be based on the structure of the solutions, where only learning methods are used to solve routing problems or combined with existing non-learning methods. In \emph{end-to-end} learning methods, either supervised or reinforcement learning is used to solve a problem from the beginning until the final solution. In \emph{hybrid} methods, learning methods are used either as a primary method to construct feasible and efficient solutions which later will be further improved with construction heuristics or learning methods facilitate to solve the inner-problems of the existing non-learning methods to solve routing problems.

\begin{figure*}
    \centering
    \includegraphics[scale=0.5]{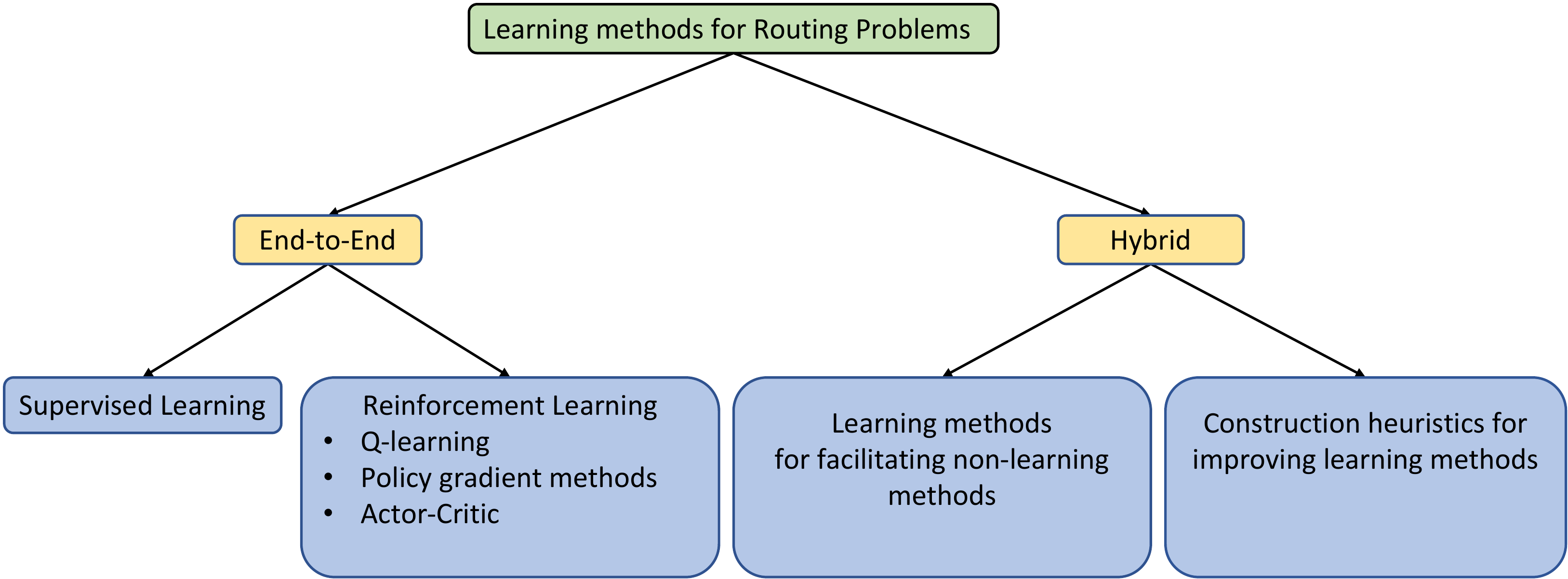}
    \caption{The taxonomy of learning methods for VRPs.}
    \label{fig:main_cat}
\end{figure*}

Another broad classification can be based on the routing problem types involving single or multiple vehicles. In principle, one can formulate the problems as single-agent reinforcement learning or multi-agent reinforcement learning problems. However, due to the shared action space among multiple vehicles, a central-controller is often used to route a fleet of heterogeneous or homogeneous vehicles. 
Figure \ref{fig:main_cat} summarizes the taxonomy of learning methods to solve VRPs. Also, since the majority of the studies focus on routing a single vehicle, we refer to the underlying problems as TSP, CVRP, VRPTW and use mTSP, mCVRP, mVRPTW to indicate multiple vehicles. 

\subsection{End-to-end Supervised Learning for VRPs}
In end-to-end models, purely learning methods construct a solution without any assistance from non-learning methods.
\emph{Supervised Learning} requires to have high-quality VRP solutions as labels that guide a model to find efficient solutions. Further, depending on how the solutions are produced, we can divide them into \emph{autoregressive} (AR) and \emph{non-autoregressive} (NAR) categories. In AR supervised learning methods, the routes are constructed step by step, one node at a time, while in NAR methods, the routes are constructed in zero-shot. 

In the case of TSP, to produce non-autoregressive solution, Held-Calp \cite{held1962dynamic} algorithm or well-known solvers such as Concorde \cite{hahsler2007tsp}, LKH3 \cite{helsgaun2017extension} is used to produce optimal solutions. Thus, for a given graph $\mathbb{G}(\mathbb{V}, \mathbb{E})$ in the two-dimensional Euclidean space, the output of the solver, $\pi^*$ is decomposed into the adjacency matrix, where for each $e_{i,j} \in \mathbb{E}$ we can compute the probability $p_{i,j}$:
\begin{align}
    p_{i,j} = 
    \begin{cases}
      1, & \text{if}  \ e_{i, j} \ \text{is in} \ \pi^* \\
      0, & \text{otherwise.}
    \end{cases}
\end{align}
Then a deep learning model with a set of parameters $\theta$, given the graph, is trained to produce an adjacency matrix with probabilities $\hat{p}_{i, j} \quad \forall e_{i, j} \in \mathbb{E}$. Then the loss is defined as a binary cross-entropy:
\begin{align}
    \mathcal{L(\theta)} = p_{i, j}\log\hat{p}_{i, j} - (1-p_{i, j})\log(1-\hat{p}_{i, j})
\end{align}
The adjacency matrix produced by the model can either follow \emph{greedy search} by picking edges with the most significant probabilities or be coupled with \emph{beam search} by preserving some number of candidate tours to be selected as the final solution.  

An exact or high-quality approximation solution $\pi^* = [v_0, \dots, v_n]$ is fed to a deep learning model to produce partial tours to maximize the conditional probability given below in autoregressive supervised learning methods:
\begin{align}
    \theta = & \argmax \log p(\pi|\pi^*, \theta) \\
    p(\pi|\pi^*, \theta) = & \prod_{i=1}^n p(v_i|v_0, \dots, v_{i-1}, \pi^*, \theta).
\end{align}

Both methods utilize advanced deep learning models such as Graph Neural Networks (GNNs) \cite{prates2019learning} and Graph Convolution Networks \cite{joshi2019efficient} to extract features of a graph and deploy  Memory Augmented Neural Networks \cite{van2020complex} and Recurrent Neural Networks (RNNs) \cite{milan2017data} to pass sequential information. Both methods require training separate sets of parameters for different graph sizes to produce near-optimal solutions for TSP \cite{joshi2019learning, joshi2021learning}. Table \ref{tb:sl} summarizes studies focused on end-to-end supervised learning.  
However, for more complex VRPs, the supervised learning method is coupled with non-learning methods, or it is fully replaced by reinforcement learning, which a priori does not require known solutions to the problems.

\begin{table*}[]
\caption{The summary of supervised learning methods with autoregressive (AR) and None-autoregressive (NAR) solutions}
\label{tb:sl}
\centering 
\begin{tabular}{@{}cccccc@{}}
\toprule
Name                                                                                            & Problems                    & AR & NAR & Loss           & Label                                                        \\ \midrule
\citet{joshi2019efficient}       & TSP &                 & \checkmark          & Cross-entropy               & Concorde                                                  \\
\citet{milan2017data}                                               & TSP                         & \checkmark               &            & Log-likelihood              & Nearest Neighbours                          \\
\citet{prates2019learning}              & a decision  TSP   &                 &            & Binary cross-entropy        & Concorde                                                              \\
\citet{joshi2019learning}                  & TSP                         & \checkmark               &            & Cross-entropy               & Concorde
\\
\citet{joshi2021learning}                                           &  TSP                           & \checkmark              &  \checkmark           &  Cross-entropy                           &     Concorde    \\                                                              
\citet{van2020complex}                                & CVRP, TSP                            &\checkmark                 &            &Cross-entropy                             &Held-Kalp algorithm, LKH-3, Concorde                                         

\\\bottomrule
\end{tabular}
\end{table*}

\subsection{End-to-end Deep Reinforcement Learning for VRPs }
While \emph{tabular} reinforcement learning methods are mostly used in the hybrid setting, in end-to-end methods, deep reinforcement learning is a popular choice due to the combinatorial nature of the action space of VRPs. We can categorize deep reinforcement learning methods applied to VRPs as value-based and policy-based.

In value-based methods such as Deep Q-learning for VRPs, neural networks are used to approximate Q-function for each state and action pair from which an optimal policy is derived. However, instead of using one-step updates as in Equation \eqref{eq:q}, n-step TD is used as a target to consider delayed rewards. Further, \emph{experience replay} is used to update Q-values with a batch of samples of the tuple $\{S_t, A_t, R_{t,{t+n}}, S_{t+n}\}$, where $R_{t,{t+n}}=\sum_{k=0}^nr_{t+k}$:
\begin{align}\label{eq:q}
    &\hat{Q}_\theta(S_t, A_t)  \gets  \hat{Q}_\theta(S_t, A_t) + \\ \nonumber
    &+ \alpha [R_{t,{t+n}}+\gamma \max_a \hat{Q}_\theta(S_{t+n}, a_{t+n}) - \hat{Q}_\theta(S_t, A_t)].
\end{align}
For instance, \cite{dai2017learning} points out that using experience replay results in fast convergence to solve combinatorial optimization problems, which is a general case when neural networks are used as approximation functions \cite{mnih2015human, riedmiller2005neural}.

Even though value-based models are generally sample-efficient as compared to the policy-gradient methods, the routing policies are complex. Therefore, in most studies, VRP policies are directly learned through policy-gradient methods. In particular,  
a parametrized policy $\pi_\theta$ is approximated using the chain rule:
\begin{align}
    \pi_\theta(a|s) = \prod_{t=0}^Tp(a_{t+1}|s_t, Y_t) \label{eq:policy}
\end{align}
where $Y_t$ is a set of visited nodes. In practice, the actor-critic structure is often used to represent two sets of neural networks to  learn $\pi_\theta$ and $V_\theta(S_0)$ respectively, where $V_\theta(S_0)$ serves as a baseline. Mean Square Error is used to update the parameters of the critic $\theta_c$ with batch size $B$ using total reward $R$:
\begin{align}\label{eq:actor}
    \theta_c \gets \theta_c + \alpha[ \frac{1}{B}\sum_{j=1}^B \nabla_{\theta_c}(R^j-\hat{V}^j_{\theta_c}(S^j_0))^2]
\end{align}
Accordingly, the actor parameters are updated using REINFORCE algorithm:
\begin{align}
    \theta_a \gets \theta_a +\alpha[\frac{1}{B}\sum_{j=1}^B\nabla_{\theta_a}\text{log}\pi_{\theta_a}(R^j-\hat{V}^j_{\theta_c}(S_0))]
\end{align}
Some studies \cite{kool2019attention} use greedy rollout of the actor with the current set of parameters to avoid the complexities of training the critic network for Equation \eqref{eq:actor}. 

To learn stochastic policy $\pi_\theta$ using an actor network, an encoder-decoder structure originated from machine translation has shown its effectiveness due to the similar nature of the problems in both fields \cite{bello2016neural}. For instance, both machine translation and VRPs aim to construct a sequence of words or nodes, given the initial set of words or nodes. The only difference is that the input set of words in machine translation also has a sequential nature. Therefore, an \emph{encoder} used in modeling VRPs differs from its original setting  in machine translation and serves as a graph embedding  responsible for passing the graph structures efficiently to a decoder \cite{nazari2018reinforcement}. 

Encoders maybe static and dynamic in nature. Static encoders embed a given initial graph only once embedding both static, $v_s$, and dynamic, $v_d$, elements of the nodes:
\begin{align}
    h^s_v = F^s(v_s), \quad 
    h^v_d = F^d(v_d) \quad \forall v \in \mathbb{V}.
\end{align}
where $F^s, F^d$ are series of  nonlinear functions and $h^s_v$ and $h^d_v$ are the embeddings of the static and dynamic elements of  node $v$.  On the other hand, dynamic encoders embed the graph at each time $t$ reflecting the changes due to the routing decisions on the dynamic parts of the nodes:
\begin{align}
    h^{d_t}_v = F^d(v_{d_t}) \quad \forall v \in \mathbb{V}.
\end{align}
The final embedding of a graph $\hat{h}_t$ is computed as the mean of embedding of all nodes in a graph at time $t$:
\begin{align}
    \hat{h}_t = \frac{1}{|\mathbb{V}|}\sum_{v \in \mathbb{V}} h^t_v
\end{align}
where, $h^t_v$ is the final embedding of node $v$. The graph embedding is the same for static encoders across all $t$. Therefore, static encoders are computed only once per problem instance, while dynamic encoders are invoked at each time step, resulting in considerable computational time. \cite{peng2019deep, xin2020step} has shown the value of dynamic encoders in improving the results to solve VRPs. 

The static and dynamic features of nodes are problem-specific and may include the coordinates of the nodes, the customer demand at each node, and others. Regardless of the types of reinforcement learning algorithms used, VRPs require to learn efficient graph representations that transform discrete and complex information between nodes and edges in a graph into a continuous vector space. 
Various models have been proposed to learn graph embeddings. For instance,  \cite{bello2016neural} uses RNNs as an encoder as a part of the Pointer Network, which \cite{nazari2018reinforcement} shows to be unnecessary due to the absence of the sequential relationship in the given initial set of nodes in a graph. Independently, \cite{da2020learning} proposed a novel graph representation called Structure2Vec that can encode both the graph and the partial solution at any time step.  \cite{kool2019attention} proposes fully attention-based encoder introducing transformers \cite{vaswani2017attention} to solve VRPs, while \cite{joshi2021learning} uses GNNs, a deep learning model dedicated to learn graph information.  
In return, other studies adopt the proposed encoders including   Pointer Networks \cite{ li2021deep},  multi-head attention \cite{ zhang2020multi, xin2020multidecoder, kim2021learning, kwon2020pomo}, recurrent neural networks (RNNs) \cite{emami2018learning}, Structure2vec \cite{james2019online, lin2021deep}  and others. However, no studies focused on investigating which models are the most efficient in graph embedding to solve VRPs.

A \emph{decoder} is a dynamic part of the actor-network invoked at each time step that incorporates the current state of a problem with the graph embedding to sample the next action. In particular, the decoder computes compatibility $u_t$ at time step $t$ that aims to extract relevant features from the inputs, which are later passed through softmax to produce the probabilities of selecting the next action:
\begin{align}
    u_t = F(\hat{h}_t, F(s_t)), \quad \pi_\theta(a|s) = \text{softmax}(u_t)
\end{align}
where, F can be a series of nonlinear transformations. 
RNNs, which enables to pass information about the sequence of the partial solution, have been a common choice for a decoder that is followed by additive attention \cite{bahdanau2014neural} to compute the alignment score between the graph embedding and the current state \cite{bello2016neural, nazari2018reinforcement,vera2019deep}. 
A fully attention-based model is another popular design choice for a decoder, where multi-head attention is used to compute the compatibility between the context node representing the state and the graph embedding \cite{kool2019attention,peng2019deep,xin2020multidecoder,falkner2020learning}. 
Table \ref{tb:rl} summarizes studies with end-to-end reinforcement learning methods. 

However, the greedy rollout resulting from a trained end-to-end reinforcement learning method may not be competitive against the hand-crafted heuristics. Therefore, it is common to use different search methods with the trained models. One of them is \emph{active search}, which aims to adopt the trained model parameters to a specifics of an instance, thus improving the solution quality  \cite{hottung2021efficient}. Another search method is \emph{random sampling}, whereby indicating the number of total solutions to be sampled, among them, we select the best solution for a reward.

\begin{table*}[]
\caption{The summary of studies with end-to-end reinforcement learning methods. CSP-Covering Salesman Problem, DVRP-dynamic VRP, PCTSP-Prize Collecting TSP, SDVRP-split delivery VRP, SPCTSP-Stochastic Prize Collecting TSP, OP-Orienteering Problem, mCVRP-multiple CVRP, mVRPSTW- multiple VRP with Soft Time Windows, MAMP-Multi-Agent Mapping Problem, MOTSP-Multi-Objective TSP, mCVRPTW-multiple CVRP with Time Windows, EVRPTW-electrical vehicle routing problem with time windows, PDP-Pickup and Delivery Problem, HCVRP-heterogeneous CVRP}
\label{tb:rl}
\centering 
\begin{tabular}{ >{\centering} p{0.15\linewidth}>{\centering}  p{0.15\linewidth}   >{\centering}p{0.15\linewidth}   >{\centering}p{0.15\linewidth}  >{\centering\arraybackslash}p{0.15\linewidth} }
\toprule[0.3pt]\midrule[0.3pt]
Study & Problems & Training Algorithm & Graph Embedding & Decoder \\ \toprule[0.3pt]\midrule[0.3pt]

\citet{bello2016neural}                                        & TSP                                                                                         & REINFORCE with baseline                 & Pointer Network                                                                                   & Pointer Network                                                                                                                                   \\\midrule[0.1pt]
\citet{james2019online}         & DVRP                                                     & REINFORCE with baseline                               & Structure2Vec                                                                & Pointer Network                                                                                                                                   \\ \midrule[0.1pt]
\citet{dai2017learning}                                     &  TSP                                                               & DQN                                          & Structure2Vec                                                                                     & -                                                                                                                                                 \\ \midrule[0.1pt]
\citet{nazari2018reinforcement}                                       & TSP, CVRP                                                                                        & REINFORCE with baseline                               & Elementwise projections                                                                                               & RNN with Attention                                                                                                                               \\\midrule[0.1pt]
\citet{xin2020step}                                         & TSP, CVRP                                             &             REINFORCE with the greedy rollout                            &  Pointer Network, Multi-head attention (Dynamic)  & Pointer Network, Multi-head attention                                                                                                              \\\midrule[0.1pt]
\citet{vera2019deep}                             & mCVRP                                                                                       &                  Actor-critic (A2C)              & RNN                                                                                               & RNN with Attention                                                 \\\midrule[0.1pt]
\citet{zhang2020multi} & mVRPSTW                                        & REINFORCE with baseline                               & Multi-head attention                                                                                      & Multi-head attention                                                                                                                                        \\\midrule[0.1pt]
\citet{kool2019attention}                                                          & TSP, CVRP, OP, PCTSP, SDVRP, SPCTSP                                                  &                  REINFORCE with baseline              & Multi-head attention                                                                                        & Multi-head attention                \\\midrule[0.1pt]
\citet{peng2019deep}  & CVRP                                                                     &                  REINFORCE with baseline              & Graph attention network                                                                           & Multi-head attention                                                                                                                                     \\\midrule[0.1pt]
\citet{sykora2020multi}                                                         &  MAMP &              REINFORCE                               &  Linear projection                                                                   &  RNN with attention                                                                                                                                                 \\\midrule[0.1pt]
\citet{xin2021multi}            & TSP, CVRP, OP, PCTSP, SDVRP, SPCTSP                                                                         &  REINFORCE with the greedy rollout               & Multi-head attention                                                                                                                 & Multi-head attention                                                                                                                                                             \\\midrule[0.1pt]
\citet{xu2021reinforcement}      & TSP, CVRP, PCTSP, SDVRP                                                          &  REINFORCE with baseline               & Multi-head attention with gate aggregation                         & Self-attention with a attentive aggregation module                                                                                            \\\midrule[0.1pt]
\citet{kim2021learning}                                   & TSP, CVRP, PCTSP                                                                             &                 REINFORCE with greedy rollout             & Multi-head attention                                                                & Multi-head attention \\\midrule[0.1pt]
\citet{kwon2020pomo}                            & TSP, CVRP                                                                                                  &  REINFORCE with baseline              & Multi-head attention                                                                                       & Multi-head attention                     \\ \midrule[0.1pt] 

\citet{emami2018learning}                                                 & TSP                                                                &                 Deterministic Policy Gradient (DPG)             & RNN                                                                                               & Sinkhorn layer                                                                                                                                    \\ \midrule[0.1pt] 
\citet{joshi2019learning}                  & TSP                                                                                         & REINFORCE  with greedy rollout & Multi-head attention  &Multi-head attention                                                        \\\midrule[0.1pt]
\citet{falkner2020learning}                & mCVRPTW                                                                                  &   REINFORCE with greedy rollout             & Self-attention, Linear projection      & Multi-head attention                                                                                                                               \\\midrule[0.1pt]
\citet{joshi2021learning}                                                       & TSP                                                                                         & REINFORCE with baseline                               & GNN                                                                                               & Multi-head attention                                                                                                                              \\\midrule[0.1pt]
\citet{li2021deep}               & CSP                                                                                        & REINFORCE 
with the greedy rollout                            & Pointer Network                                                                                    & RNN with attention                                                                                                                                \\\midrule[0.1pt]
 \citet{drori2020learning} & TSP, VRP  & REINFORCE with baseline & Graph Attention Network (GAT) & Multi-head attention
                                                                                                \\ \midrule[0.1pt]
 \citet{lin2021deep} & EVRPTW  & REINFORCE with greedy rollout  & Structure2Vec & RNN with Attention  
 \\ \midrule[0.1pt]
 
  \citet{li2020deep} & MOTSP  & Actor-critic (A2C) & 1D-Conv & GRU with Attention  
 \\\midrule[0.1pt]
 
 \citet{li2021heterogeneous} & PDP  & REINFORCE with greedy rollout & Multi-head attention & Multi-head attention  
 \\\midrule[0.1pt]
 
 \citet{licvrp2021deep}& HCVRP  & REINFORCE with greedy rollout & Multi-head attention & Feed-Forward networks, Multi-head attention 
 \\ \midrule[0.1pt]
 
  \citet{duan2020efficiently}& CVRP  & REINFORCE with greedy rollout, SUPERVISE with policy-sampling  & Graph Convolutional Networks & GRU with context-based attention, Multi-layer Perceptron 
 \\

 \bottomrule
\end{tabular}
\end{table*}
\section{Hybrid Methods}\label{sec:learnhybrid}
Hybrid Methods aim to take advantage of both learning and non-learning methods to solve challenging VRPs. Based on the structure of the hybrid method solutions, we can categorize them into two groups. In the first group, learning methods serve as a supporting tool to address the internal subproblems of non-learning methods. In the second group, learning methods are a primary tool to produce a solution, which is improved with construction heuristics. 
\subsection{Learning Methods for Facilitating non-learning methods}
Combining Q-learning with meta-heuristics dates back to \cite{gambardella1995ant}, which proposed to use ant-colony in a distributed setting to explore solutions for TSP. In particular, each ant represents an agent to construct solutions and fill a Q-table associated with moving from one city to another. At the same time, the reward function reinforces selecting tours with short lengths. Later, \cite{dorigo1997ant} further developed the idea of enhancing cooperation between ants/agents through the introduction of updated pheromone values assigned to each edge in the graph. The study has shown that combining the ant-colony system with local search achieves competitive results in solving TSP. 

\cite{liu2009study} combines the genetic algorithm with reinforcement learning to solve TSP, where the Q-learning algorithm is modified along with the mutation operation to produce efficient tours, which later can be enforced with the local search.
Similarly, \cite{alipour2018hybrid} combines the genetic algorithm with multi-agent reinforcement learning, where MARL is used to generate an
initial solution which later will be improved by a genetic algorithm. 

\cite{hottung2019neural} presents a model to use reinforcement learning for neighborhood search to solve CVRP and split delivery VRP (SDVRP). In particular, a feasible solution is destroyed by a destroy operator in several locations. Then a repair operator is trained using RL to connect the partial solutions. In particular, an encoder-decoder model is proposed to learn the embeddings of the end nodes of the partial solutions and produce the probabilities of connecting nodes to construct a complete solution.  The results outperform \cite{kool2019attention, nazari2018reinforcement} for VRP and SDVRP for different node sizes and show competitive results with LKH3, unified hybrid genetic search (UHGS).  Also, \cite{syed2019neural} proposes using \cite{nazari2018reinforcement} model combined with Large Neighborhood Search (LNS) to solve VRPWT in ride-hailing services. They train the neural networks using supervised learning for the insertion operation inside LNS. 
\cite{silva2019reinforcement} further enhances the idea of \cite{fernandes2009multiagent} using multi-agent systems to learn which meta-heuristic to apply to solve VRPTW and develops Adaptive Local Search through Q-learning. Q-learning aims to determine the sequence of neighborhoods that need to be explored first in order to maximize objective function by picking the actions presented as different operations available from Variable Neighborhood Search. 

Along with approximated dynamic programming methods used to solve VRPs \cite{ulmer2019offline, ulmer2018budgeting}, the combination of reinforcement learning with dynamic programming is an emerging research direction. For instance, 
\cite{cappart2021combining} proposes to combine reinforcement learning with constraint programming (CP) using a dynamic programming approach to solve TSP, VRP, and TSPTW. Deep Q-learning and Proximal Policy Optimization (PPO) are integrated into CP search algorithms. 
\cite{kool2021deep} aims to mitigate the curse of dimensionality present in dynamic programming by restricting its search space to policies produced by learning methods. Identical to \cite{joshi2019efficient}, the study generates the heatmap for the edges of a given graph that indicate the probabilities of entering those edges into the solution. Then this heatmap is used for branching in forwarding Dynamic Programming. 
\cite{xu2020deep, yang2018boosting} uses neural networks to approximate the value functions for dynamic programming, which expedites the solution time. 
\cite{delarue2020reinforcement} a policy iteration algorithm to solve CVRP. The authors propose a simple model consisting of fully connected hidden layers and a ReLU activation function to estimate the value of each state defined as the set of unvisited nodes. Then for policy evaluation, they used Mixed Integer Programming (MIP) to select the actions, combining the learned value functions. The resulting MIP is solved using the branch-and-cut approach of \cite{anderson2020strong}. 

Another set of studies uses learning methods to pick heuristics to solve routing problems. For instance, \cite{ma2021learning, wu2021learning} aim to learn improved heuristics. They embed the nodes and positions in a given solution separately, wherein in the middle of the encoding, they are shared with each other, but at the end, the encoder produces different node features and position feature embedding that are passed to the decoder. The decoder produces the matrix of probabilities to select the pair of nodes to use for local operation. \cite{chen2019learning} proposes NeuRewriter model, that given an initial solution, picks the region to be modified and applies some learned rewriting rules to improve the solution. Both region-picker and rule-picker policies are trained using reinforcement learning to solve TSP and CVRP. Similarly, \cite{gao2020learn} proposes to learn a general heuristic to solve VRP, VRPWT that given an initial solution destroys some part of the solution and learns to reconstruct an improved version of it. The study modify GNNs to embed both the nodes and edges, which passed through GRU to define the order of the nodes of the new repaired part for the solution. 
Also, \cite{li2021learning} proposes to delegate subproblems of VRPs to black-box solvers and trains a subproblem generator model based on transformers and supervised learning. In particular, they restrict the possible number of subproblems to solve given an initial solution by looking at only subsets of the visited nodes and restricting the number of routes to be considered by utilizing spatial locality.

\subsection{Improving Learning Methods with Heuristics}

Along with sampling methods such as beam search, active search and random sampling, trained learning methods can be combined with traditional construction heuristics. In particular, a greedy solution produced by a learning method is used as an initial solution. For instance, \cite{deudon2018learning} solves TSP with the MHA-based encoder and the Pointer Network decoder along with a 2-opt local search. For solving VRP and VRPTW on a large scale, \cite{zhao2020hybrid} used \cite{nazari2018reinforcement}' model with an adaptive critic later combined with local search algorithms such as OR-tools and LNS. Also,  
\cite{ma2019combinatorial} introduces Graph Pointer Nets (GPNs) to solve TSP, where a graph encoder made of GNNs is combined with Pointer Networks. Then the paper proposes a Hierarchical RL to solve TSPTW, where each layer of the neural network learns to solve TSP with some constraint. By combining GPNs with local search, they obtained comparable results with the construction heuristics. They also search real and random data instances of TSPWT and demonstrate comparable results with the construction heuristics. Another study by
\cite{zhang2020deep} solves TSP with Time Windows and Rejections, where the objective is to minimize the rejection and distance costs. The paper proposes to use the AM model to solve the regular TSP and deploy a heuristic that will check the feasibility of the produced solution according to Time Windows and determine if an order has been rejected. This will help compute the rewards function correctly according to TSPWTR and update the parameters of NNs using the baseline rollout.

Learning methods can also be combined with prescriptive methods to solve complex routing problems. For instance, \cite{Lin2019EfficientCM}
discusses the problem of managing a large set of available homogeneous vehicles for online ride-sharing platforms. In this work, the authors propose Contextual DQN and Contextual Actor-Critic networks to learn a  state-value function shared by all agents representing each vehicle. The learned state-value functions are used for efficient allocation solved by linear programming.

Another set of studies combines learning methods with Monte-Carlo Tree Search. For instance, \cite{fu2021generalize} proposes the idea of training TSP in a supervised manner in small instances using LKH as the optimal solution and then using the trained model to solve large instances. First, the authors train a graph convolutional residual network with an attention mechanism on a graph with $m$ nodes. Then they take a larger graph and split it into several graphs with size $m$, where each node can be present in several graphs. Then they use the trained model for each subgraph to generate the heatmap that shows the probabilities of connecting nodes in the graph. To combine the heatmap and produce a single solution, they sum the probabilities for each arc computed across all the subgraphs, and this summed heatmap is used for Monte-Carlo Tree Search to improve further the solution, similar to \cite{xing2020graph}.  The study solves TSPs with up to 10,000 nodes in a reasonable time.

\begin{table*}[]
\centering
\caption{The summary of studies with hybrid methods. CVRP-Capacitated VRP, CVRPTW-CVRP with Time Windows, MARL-Multi-agent Reinforcement Learning, SDVRP-Split Delivery VRP, TSPTW-TSP with Time Windows, TSPTWR-TSP with Time Windows and Rejection, VRPTW-VRP with Time Windows, SDDPVD-same-day delivery problem with vehicles and drones, mVRPTW- multiple VRP with Time Windows,  DVRP-dynamic VRP, ATSP-asymmetric TSP}
\label{tb:hybrid}
\begin{tabular}{ >{\centering}p{0.2\linewidth}  >{\centering}p{0.15\linewidth}  >{\centering}p{0.1\linewidth}  >{\centering}p{0.15\linewidth} >{\centering\arraybackslash}p{0.15\linewidth}   }
\toprule[0.3pt]\midrule[0.3pt]

Name                                                                                                                                 & Problems                             & Type   & Learning Method                                     & Non-learning method                                                                                                                                    \\ \toprule[0.3pt]\midrule[0.3pt]
\citet{dorigo1997ant}                                                 & TSP                                  & First         & Q-Learning                                                    & Ant-colony                                                                                                                                                       \\\midrule[0.1pt]
\citet{liu2009study}                                                              & TSP                                  & First         & Q-learning                                                    & Genetic Algorithm                                                                                                                                                \\\midrule[0.1pt]
\citet{alipour2018hybrid} & TSP                                  & First  & MARL                                                          & Genetic Algorithm,  2-opt                                                       \\\midrule[0.1pt]
\citet{hottung2019neural}                                                         & CVRP, SDVRP                       & First         & Policy-based RL                                               & Neighborhood Search                                                                                                                                              \\\midrule[0.1pt]
\citet{syed2019neural} & VRPTW in ride-hailing services       & First         & Supervised Learning                                                           & Large Neighborhood Search                                                                                                                                        \\\midrule[0.1pt]
\citet{fernandes2009multiagent}                                    & VRPTW                                & First         & Q-learning                                                    & Adaptive Local Search                                                                                                                                             \\\midrule[0.1pt]
\citet{cappart2021combining}                                          & TSPTW                   & First         & DQN, Proximal Policy Optimization (PPO)                                                 & Constraint Programming, Dynamic Programming \\\midrule[0.1pt]
\citet{kool2021deep}                                                                        & TSP, CVRP, TSPTW                      & First         & Supervised learning                                                           & Dynamic Programming                                                                                                                                              \\\midrule[0.1pt]
\citet{xu2020deep}                                                  & TSP & First         & Unsupervised learning                                                           & Dynamic Programming                                                                                                                                              \\\midrule[0.1pt]
\citet{delarue2020reinforcement}                                                 & CVRP                                 & First         & Value-based RL                                                            & Mixed Integer Programming   (MIP)                                                                                                                                                   \\\midrule[0.1pt]
\citet{ma2021learning}                                            & CVRP                                  & First         & The actor-critic variant of PPO                                               & 2-opt, swap and insert                                                                                                                                           \\\midrule[0.1pt]
\citet{chen2019learning}  & CVRP & First         & Actor-Critic   & Halide rewriter \cite{ragan2013halide}               \\\midrule[0.1pt]
\citet{li2021learning}                                                                               & Large-scale CVRP                      & First         & Supervised Learning                                                  & MIP solver                                                                                                                                                       \\\midrule[0.1pt]
\citet{deudon2018learning}                                                                                           & TSP                                  & Second           & MHA-based encoder, Transformers Decoder                       & 2-opt local search                                                                                                                                               \\\midrule[0.1pt]
\citet{zhao2020hybrid}                                                                                               & CVRP and VRPTW in large scale         & Second           & Policy Gradient                                               & OR-tools and LNS                                                                                                                                                 \\\midrule[0.1pt]
\citet{ma2019combinatorial}                                                                                          & TSPTW                                & Second           & Hierarchical Policy based RL                                  & Local search                                                                                                                                                     \\\midrule[0.1pt]
\citet{zhang2020deep}                                                                                                & TSPTWR                               & Second           & Policy based RL                                               & Tabu search                                                                                                                                             \\\midrule[0.1pt]
\citet{Lin2019EfficientCM}                                                                                           & Large-scale fleet management problem & Second           & Contextual DQN and Contextual Actor-Critic                    & Linear programming                                                                                                                                               \\\midrule[0.1pt]

\citet{chen2022deep}                                                                                             & SDDPVD                                  & First           & Deep Q-learning & A policy function approximation                                                                                                                                        \\\midrule[0.1pt]
\citet{tyasnurita2017learning}  & Open VRP & First
& Hyper-heuristic classification & Modified Choice Function All Moves\\\midrule[0.1pt]%
\citet{xing2020graph} & TSP & Second & Supervised Learning to learn heat maps & Monte Carlo Tree Search \\\midrule[0.1pt]
\citet{fu2021generalize}  & TSP & Second & Supervised Learning to learn heat maps & Monte Carlo Tree Search \\\midrule[0.1pt]
\citet{hottung2020learning} & TSP, CVRP & Second & Supervised Learning to learn latent space & Unconstrained continuous optimization \\\midrule[0.1pt]

\citet{gutierrez2019selecting} & mVRPTW & Second & Supervised Learning & HMOEA-06 \cite{tan2006hybrid}, MOGA-06 \cite{ombuki2006multi}, MMOEAD-15 \cite{qi2015decomposition}, HMPSO-16 \cite{wu2016vehicle} \\\midrule[0.1pt]

\citet{da2020learning} & TSP & Second & DRL, Policy Gradient & 2-opt local search  \\\midrule[0.1pt]

\citet{fu2019targeted} & TSP & First & MCTS with reinforcement learning & 2-opt local search \\\midrule[0.1pt]

\citet{wu2021learning} & TSP, CVRP & First & Policy gradient & Pairwise local operators \\\midrule[0.1pt]

 \citet{xin2021neurolkh} & TSP, CVRP, CVRPTW & First & Supervised Learning & LKH Algorithm \\\midrule[0.1pt]
\citet{yang2018boosting} & TSP & First & Deep Q-Learning & Dynamic Programming
\\\midrule[0.1pt]
\citet{gao2020learn} & CVRP, CVRPTW & Second & Actor-Critic &  Very Large-scale Neighborhood Search \\\midrule[0.1pt]
\citet{joe2020deep} & DVRP & First &  TD learning  & Simulated Annealing \\\midrule[0.1pt]
\citet{gambardella1995ant} & TSP, ATSP & First  & Q-learning & Ant System \\

\bottomrule
\end{tabular}
\end{table*}

\section{Single vs. Multiple VRP Formulations}\label{sec:forms}
The majority of VRPs in practice involve routing either multiple heterogeneous or homogeneous vehicles \cite{sun2019feasible}. However, the straightforward application of learning methods to such VRPs is challenging due to the presence of several vehicles in a graph that all can influence the shared environment. This section presents the MDP formulations for routing a single vehicle using CVRP as an example. In addition to the formulation, we present the comparison results for TSP, CVRP and VRPTW. Finally, we discuss several approaches present in the literature to formulate multiple VRPs.  

\subsection{A Single Vehicle Routing Problems}

A single vehicle routing problem is defined in a graph $\mathbb{G}(\mathbb{V}, \mathbb{E})$ consisting of a set of nodes $\mathbb{V} = \{v_0, \dots, v_n \}$ and a set of edges $\mathbb{E} = \{(v_i, v_j): i < j, v_i, v_j \in \mathbb{V} \}$, where each node represents  depot, $v_0$, or customer's locations, $\mathbb{V} \setminus v_0$. We aim to find a sequence of nodes to be visited by a vehicle to minimize the total costs defined as the total distance travelled. %
We can further add capacity constraints by setting the maximum load, $l_{\text{max}}$, to be carried by a vehicle and let $d_v$ represent a customer demand at node $v$. The problem naturally fits a single-agent reinforcement learning, where a vehicle represents the agent, and the environment is a simulated CVRP. Then the agent is trained to select action $A_{t+1} \in \mathbb{V}$ at each time step $t$, representing the node index to be visited next given the current state $S_t$. In CVRP with a single vehicle, the current state is defined by a tuple $S_t=\{a_t, l_t, Y_t\}$, representing the current location of a vehicle, the remaining load, and the set of served customers. This state representation allows to fully capture the dynamics of the problem and make the future decisions based only on the current state, thus preserving the Markov property.  At each time step $t$, we can update the remaining load as follows:
\begin{align*}
    l_{t+1} = l_{\text{max}} \quad \text{if}\ a_t=v_0 \quad \text{and} \quad  l_{t+1} = l_{t}-d_{a_t} \quad  \text{otherwise.}
\end{align*}
Similarly, set $Y_t$ includes all served customers with $d^t_v=0$ for all $v \in \mathbb{V} $, which is updated based on the decision of the agent at time $t$:
\begin{align*}
    d^{t+1}_v = d^{t}_v \quad  \text{if}\ a_t\neq v
    \quad \text{and} \quad
    d^{t+1}_v =d^{t}_v-l_{t} \quad \text{otherwise.}
\end{align*}
For each decision at time $t$, we calculate intermediate reward $r_t$ defined as the distance between the current location, $a_t$ and to be visited node $a_{t+1}$ from which we can derive the total reward, R:
\begin{align}
    R = \sum_{t=0}^T r_t = \sum_{t=0}^{T-1} c_{a_{t}, a_{t+1}}
\end{align}
The majority of the studies focus on routing a single vehicle with some modification to the above CVRP by considering time window constraints \cite{fernandes2009multiagent, zhao2020hybrid, xin2021neurolkh}, allowing split deliveries \cite{nazari2018reinforcement, hottung2019neural}, introducing dynamic demand \cite{james2019online} or generalizing to various forms of TSP \cite{kool2019attention, xin2020multidecoder, ma2019combinatorial, zhang2020deep}. 
However, comparing the proposed models with other learning-based or non-learning methods is a challenging task due to the absence of widely-accepted benchmark instances and differences in the hardware settings. Also, most of the studies, except a few, using only randomly generated instances, which prevents to fairly compare them with non-learning methods that have been widely tested on real-world datasets. 

Tables \ref{tb:tsp} and \ref{tb:cvrp} present the comparison of the surveyed studies to solve Euclidean distance TSP and CVRP, respectively, on randomly generated graphs with 20, 50, and 100 nodes. The results include the averages of objective values, gaps, and computational time for 10,000 instances if not indicated otherwise. The gaps are measured against the best-performing method. 

We compare end-to-end learning methods \cite{ kool2019attention, kim2021learning,  kwon2020pomo,  joshi2019efficient, xin2021multi,  dai2017learning}, hybrid methods \cite{ma2021learning,wu2021learning,da2021learning,kool2021deep, hottung2020learning,deudon2018learning, hottung2019neural} in their different configurations and used Concorde, LKH and OR-tools as baselines. We present the results reported in the original studies. 
The hybrid model of \cite{ma2021learning} performs slightly better both to solve TSP and CVRP than the rest of presented studies. However, it is extremely challenging to compare the computational times due to the differences in the hardware types and quantities and the number of batches used for inference. Also, \cite{kim2021learning} placed a time budget in the experiments, resulting in small running times. Tables \ref{tb:tsplib} and \ref{tb:cvrplib} show the generalization capacity of \cite{kool2019attention, kwon2020pomo, wu2021learning, kim2021learning, ma2021learning} on well-known TSPLib \cite{reinelt1991tsplib} and CVRPLib \cite{uchoa2017new} datasets. We report the results of \cite{kwon2020pomo, kool2019attention}
based on \cite{ma2021learning},  the results of \cite{da2021learning} are based on \cite{kim2021learning} and we report the results of  \cite{ma2021learning, wu2021learning, kim2021learning} from the original studies. \cite{ma2019combinatorial} performs the best across both datasets. 

\begin{table*}[]
\caption{The performance comparison to solve TSP on random instances.}
\label{tb:tsp}
\centering 
\begin{threeparttable}
\begin{tabular}{@{}cccccccccc@{}}
\toprule
                                          & \multicolumn{3}{c}{TSP20}       & \multicolumn{3}{c}{TSP50}       & \multicolumn{3}{c}{TSP100}        \\ 
\cmidrule(lr){2-4} \cmidrule(lr){5-7} \cmidrule(lr){8-10}
Method                                    & Obj.  & Gap      & Time        & Obj.  & Gap      & Time        & Obj.  & Gap      & Time          \\ \midrule
Concorde                                  & 3.83  & -        & 3m        & 5.70   & -        & 10m       & 7.76  & -        & 1h          \\
LKH                                       & 3.83  & 0.00\%   & 38s       & 5.70   & 0.00\%   & 5m        & 7.76  & 0.00\%   & 20m         \\
OR-Tools                                  & 3.86  & 0.94\%   & 42s       & 5.85  & 2.87\%   & 5m        & 8.06  & 3.86\%   & 23m         \\ \midrule
\citet{joshi2019efficient}, GCN-BS                                    & 3.84\tnote{1} & 0.01\%   & 12m       & 5.70   & 0.01\%   & 18m       & 7.87  & 1.39\%   & 40m         \\
\citet{dai2017learning}                                   & 3.89  & 1.42\%   &    -         & 5.99  & 5.16\%   &       -      & 8.31  & 7.03\%   &      -         \\
\citet{kool2019attention}, AM(N=1,280)                               & 3.83  & 0.06\%   & 14m         & 5.72  & 0.48\%   & 47m         & 7.94  & 2.32\%   & 1.5h          \\
\citet{kool2019attention}, AM(N=5,000)                               & 3.83  & 0.04\%   & 47m         & 5.72  & 0.47\%   & 2h          & 7.93  & 2.18\%   & 5.5h          \\
\citet{kim2021learning}, AM+LCP \{640,10\}                         & 3.84\tnote{1}  &  \textbf{0.00\% }  & 0.18s       & 5.70   & 0.13\%   & 0.30s       & 7.86  & 1.25\%   & 0.57s         \\
\citet{kim2021learning}, AM+LCP \{1280,10\}                        & - & - & -          & 5.70   & 0.10\%   & 0.45s       & 7.85  & 1.13\%   & 0.90s         \\
\citet{kim2021learning}, AM+LCP*\{1280,45\}                        & - & - & -          & 5.70   & 0.02\%   & 2.48s       & 7.81  & 0.54\%   & 4.30s         \\
\citet{kwon2020pomo}, POMO                                      & 3.83  & 0.04\%   & 1s        & 5.70\tnote{1} & 0.21\%   & 2s        & 7.80   & 0.46\%   & 11s         \\
\citet{kwon2020pomo}, POMO$\times$8 augment                            & 3.83  &  \textbf{0.00\% }    & 3s        & 5.69\tnote{1} & 0.03\%   & 16s       & 7.78  & 0.15\%   & 1m          \\
\citet{xin2021multi}, MDAM-BS                                   & 3.84\tnote{1} &  \textbf{0.00\% }   & 3m        & 5.70   & 0.03\%   & 14m       & 7.79  & 0.38\%   & 44m         \\ \midrule
\citet{kool2021deep}, DPDP (100k)                               & -     & -        & -           & -     & -        & -           & 7.77\tnote{1} & 0.00\%   & 3h          \\
\citet{ma2021learning}, DACT (T=1k)                               & 3.83  & 0.04\%   &24s & 5.70   & 0.14\%   & 1m & 7.89  & 1.62\%   & 4m   \\
\citet{ma2021learning}, DACT(T=5k)                                & 3.83  &  \textbf{0.00\% }   & 2m & 5.70   & 0.02\%   & 6m  & 7.81  & 0.61\%   & 18m   \\
\citet{ma2021learning}, DACT(T=10k)                               & 3.83  &  \textbf{0.00\% }    &5m  & 5.70   & 0.01\%   & 13m & 7.79  & 0.37\%   & 40m   \\
\citet{ma2021learning}, DACT$\times$4 augment                       & 3.83  &  \textbf{0.00\% }    & 10m & 5.70   & \textbf{0.00\% }   & 1h & 7.77  & \textbf{0.09\% }  & 2.5h \\
\citet{wu2021learning}, (T=1,000) & 3.83  & 0.03\%   & 12m         & 5.74  & 0.83\%   & 16m         & 8.01  & 3.24\%   & 25m           \\
\citet{wu2021learning}, (T=3,000) & 3.83  &  \textbf{0.00\% }         & 39m         & 5.71  & 0.34\%   & 45m         & 7.91  & 1.85\%   & 1.5h          \\
\citet{wu2021learning}, (T=5,000) & 3.83  &  \textbf{0.00\% }         & 1h          & 5.70   & 0.20\%    & 1.5h        & 7.87  & 1.42\%   & 2h            \\
\citet{deudon2018learning}, EAN \{M:1280\}                            & 3.84  & 0.11\%   &    -         & 5.77  & 1.28\%   &     -        & 8.75  & 12.70\%  &       -        \\
\citet{deudon2018learning}, 2OPT \{M:1280\}                       & 3.84  & 0.09\%   &     -        & 5.75  & 1.00\%   &       -      & 8.12  & 4.64\%   &      -         \\
\citet{hottung2020learning}                               & -     &  \textbf{0.00\%  \tnote{2}} & 11m\tnote{2}       & -     & 0.02\%\tnote{2} & 22m\tnote{2}       & -     & 0.34\% & 55m\tnote{2}         \\
\citet{da2021learning}                              & 3.84\tnote{1} &  \textbf{0.00\% }    & 15m       & 5.70   & 0.12\%   & 29m       & 7.83  & 0.87\%   & 41m          \\

\bottomrule
\end{tabular}
\begin{tablenotes}\footnotesize
\item[1] 
the obj. values obtained by Concorde or LKH may be slightly different from DACT since the 10,000 instances are randomly generated.
\item[2] the obj. values, gaps, or time are obtained based on 2,000 instances in their original papers
\end{tablenotes}
\end{threeparttable}
\end{table*}

\begin{table*}[]
\caption{The performance comparison to solve CVRP on random instances.}
\label{tb:cvrp}
\centering 
\begin{threeparttable}
\begin{tabular}{@{}cccccccccc@{}}
\toprule
 & \multicolumn{3}{c}{CVRP20}      & \multicolumn{3}{c}{CVRP50}       & \multicolumn{3}{c}{CVRP100}       \\ 
 \cmidrule(lr){2-4} \cmidrule(lr){5-7} \cmidrule(lr){8-10}
Method                                    & Cost   & Gap     & Time         & Cost    & Gap    & Time          & Cost    & Gap    & Time           \\ \midrule

LKH3                                   & 6.14   & 0.00\%  & 1h        & 10.38   & 0.00\% & 4h         & 15.68   & 0.00\% & 8h          \\
OR-Tools                              & 6.46   & 5.68\%  & 2m            & 11.27   & 8.61\% & 13m             & 17.12   & 9.54\% & 46m              \\ \midrule
\citet{nazari2018reinforcement}\{M: 10\}                               & 6.40    & 4.31\%  & -            & 11.15   & 7.46\% & -             & 16.96   & 8.18\% & -              \\

\citet{kool2019attention}, AM\{M: 1280\}                             & 6.25   & 1.90\%  & 0.05s        & 10.62   & 2.40\% & 0.14s         & 16.23   & 3.52\% & 0.34s          \\
\citet{kool2019attention}, AM\{M: 2560\}                             & 6.25   & 1.80\%  & 0.06s        & 10.61   & 2.24\% & 0.31s         & 16.17   & 3.14\% & 0.75s          \\
\citet{kool2019attention}, AM\{M: 7500\}                             & 6.24   & 1.65\%  & 0.09s        & 10.59   & 2.06\% & 0.36s         & 16.14   & 2.91\% & 1.42s          \\
\citet{kim2021learning} \{640,1\}                          & 6.17   & 1.15\%  & 0.07s        & 10.56   & 1.74\% & 0.15s         & 16.05   & 2.58\% & 0.30s          \\
\citet{kim2021learning} \{1280,1\}                         & 6.16   & 0.92\%  & 0.09s        & 10.54   & 1.54\% & 0.20s         & 16.03   & 2.43\% & 0.45s          \\
\citet{kim2021learning} \{2560,1\}                         & 6.15   & 0.84\%  & 0.14s        & 10.52   & 1.38\% & 0.31s         & 16.00      & 2.24\% & 0.77s          \\
\citet{kim2021learning} \{6500,1\}                         & -&-&-         & -&-&-             & 15.98   & 2.11\% & 1.73s          \\
\citet{kwon2020pomo}                             & 6.17\tnote{1}  & 0.82\%  & 1s         & 10.49   & 1.14\% & 4s          & 15.83   & 0.98\% & 19s          \\
\citet{kwon2020pomo}$\times$8 augment                    & 6.14\tnote{1}  & 0.21\%  & 5s         & 10.42   & 0.45\% & 26s         & 15.73   & 0.32\% & 2m           \\ \midrule
\citet{hottung2019neural}\{I: 2000\}                           & 6.19   & 0.88\%  & 1.00s        & 10.54   & 1.54\% & 1.63s         & 16.00      & 1.97\% & 2.18s          \\
\citet{kool2021deep} (100k)                      & -      & -       & -            & -       & -      & -             & 15.69\tnote{1}  & 0.31\% & 6h           \\
\citet{ma2021learning} (T=1k)                               & 6.15   & 0.28\%  & 33s & 10.61   & 2.13\% & 2m   & 16.17   & 3.18\% & 5m     \\
\citet{ma2021learning} (T=5k)                               & 6.13   & 0.00\%  & 3m   & 10.48   & 1.01\% & 8m    & 15.92   & 1.55\% & 23m    \\
\citet{ma2021learning} (T=10k)                              & 6.13   & -0.04\% & 6m   & 10.46   & 0.79\% & 16m   & 15.85   & 1.12\% & 45m   \\
\citet{ma2021learning}$\times$6 augment                            & 6.13   & \textbf{-0.08\%} & 35m & 10.39   & \textbf{0.14\%} & 1.5h & 15.71   & \textbf{0.19\%} & 4.5h \\
\citet{wu2021learning} (T=1,000) & 6.16\tnote{1}   & 0.48\%  & 23m          & 10.71   & 3.16\% & 48m           & 16.30    & 3.89\% & 1h             \\
\citet{wu2021learning} (T=3,000) & 6.14\tnote{1}   & 0.19\%  & 1h           & 10.55   & 1.65\% & 2h            & 16.11   & 2.73\% & 3h             \\
\citet{wu2021learning} (T=5,000) & 6.12\tnote{1}   & -0.03\%  & 2h           & 10.45   & 0.70\% & 4h            & 16.03   & 2.21\% & 5h             \\

\citet{hottung2020learning}                     & 6.14\tnote{2} & -       & 21m        & 10.40\tnote{2} & -      & 41m\tnote{2}         & 15.75\tnote{2} & -      & 1.5h\tnote{2}         \\

\bottomrule
\end{tabular}
\begin{tablenotes}\footnotesize
\item[1] 
the obj. values obtained by Concorde or LKH may be slightly different from DACT since the 10,000 instances are randomly generated.
\item[2] the obj. values, gaps, or time are obtained based on 2,000 instances in their original papers
\end{tablenotes}
\end{threeparttable}
\end{table*}

Solving large-scale routing problems using learning methods is still in development. One of such studies \cite{li2021learning} presents the hybrid model that outperforms the hand-crafted heuristics such as Random, Count-based, and Max-min. Table \ref{tb:cvrplarge} reported from \cite{li2021learning} shows the comparison with LKH, OR-tools and learning-based studies \cite{kool2019attention,  chen2019learning}. Even though there have been some studies to solve VRPTW \cite{zhao2020hybrid, fernandes2009multiagent, zhang2020multi}, the settings of the problem are different for each study, which prevents the comparisons between them. Therefore, in  
Table \ref{tb:vrptw} we report the comparison between  \cite{falkner2020learning} and \cite{kool2019attention}  on the well-known Solomon dataset \cite{solomon1987algorithms}. 

\begin{table*}[]
\centering
\caption{The performance comparison to solve large-scale CVRP on random instances.}
\label{tb:cvrplarge}
\begin{tabular}{@{}ccccccc@{}}
\toprule
& \multicolumn{2}{c}{CVRP500}      & \multicolumn{2}{c}{CVRP1000}       & \multicolumn{2}{c}{CVRP2000}       \\ 
 \cmidrule(lr){2-3} \cmidrule(lr){4-5} \cmidrule(lr){6-7}
Method                       & Cost      & Time          & Cost     & Time            & Cost     & Time            \\ \midrule
LKH3 (95\%)                 & 62        & 4.4min        & 120.02   & 18min           & 234.89   & 52min           \\
LKH3 (30k)                  & 61.87     & 30min         & 119.88   & 77min           & 234.65   & 149min          \\
OR-Tools                     & 65.59     & 15min         & 126.52   & 15min           & 244.65   & 15min           \\
\citet{kool2019attention}, AM sampling                  & 69.08     & 4.70s         & 151.01   & 17.40s          & 356.69   & 32.29s          \\
\citet{kool2019attention}, AM greedy                    & 68.58     & 25ms          & 142.84   & 56ms            & 307.86   & 147ms           \\
\citet{chen2019learning}                  & 73.6      & 58s           & 136.29   & 2.3min          & 257.61   & 8.1min          \\
Random                       & 61.99     & 71s     & 120.02   & 3.2min    & 234.88   & 6.4min    \\
Count-based                  & 61.99     & 59s     & 120.02   & 2.1min    & 234.88   & 5.3min     \\
Max Min                      & 61.99     & 59s    & 120.02   & 2.5min   & 234.89   & 5.2min    \\
\citet{li2021learning} (Short) & 61.99     & 38s     & 119.87   & 1.5min    & 234.89   & 3.4min     \\
\citet{li2021learning} (Long)  & \textbf{61.7}      & 76s           & \textbf{119.55}   & 3.0min          & \textbf{233.86}   & 6.8min          \\ \bottomrule
\end{tabular}
\end{table*}

\begin{table*}[]
\centering
\caption{The performance comparison to solve VRPTW  on the Solomon benchmark problems.}
\label{tb:vrptw}
\begin{tabular}{@{}cccccccccc@{}}
\toprule
                      & Model         & \multicolumn{4}{c}{VRPTW20}         & \multicolumn{4}{c}{VRPTW50}            \\
                      \cmidrule(lr){3-6} \cmidrule(lr){7-10}
                      &               & Cost    & K    & Dist    & Tinf  & Cost     & K     & Dist    & Tinf   \\
                      \midrule
\multirow{10}{*}{TW1} & \citet{falkner2020learning}, OR-Tools-AU       & 2577.08 & 4.18 & 643.77  & 0.22s & 4344.88  & 6.30   & 1110.68 & 1.76s  \\
                      & \citet{falkner2020learning}, OR-Tools-GLS      & 2522.85 & 4.11 & 617.77  & 8.00s & 4213.23  & 6.15  & 1090.36 & 8.06s  \\
                      & random (1000) & 3036.39 & 5.68 & 1200.37 & -     & 7297.53  & 11.80  & 2914.86 & -      \\
                      & \citet{kool2019attention}, AM+TW(greedy) & 3766.88 & 5.29 & 1264.40  & 0.05s & 7189.43  & 9.42  & 2882.53 & 0.12s  \\
                      & \citet{kool2019attention}, AM+TW(sampl.) & 3041.24 & 5.74 & 1202.64 & 0.12s & 7327.09  & 11.92 & 2921.39 & 0.38s  \\
                      & \citet{kool2019attention}, AM+TW(t10240) & 2750.06 & 5.27 & 1163.58 & 0.95s & 6878.84  & 11.28 & 2865.30  & 3.08s  \\
                      & \citet{falkner2020learning}, JAMPR(greedy) & 1862.40  & 2.25 & 966.74  & 0.10s & 3055.94  & 5.42  & 1733.24 & 0.24s  \\
                      & \citet{falkner2020learning}, JAMPR(sampl.) & \textbf{1716.60}  & 2.29 & 965.42  & 0.86s & \textbf{2691.55}  & 4.03  & 1811.06 & 3.07s  \\ \midrule
\multirow{10}{*}{TW2} & \citet{falkner2020learning}, OR-Tools-AU       & 635.06  & 4.23 & 635.06  & 0.37s & 1123.82  & 6.72  & 1123.82 & 3.81s  \\
                      & \citet{falkner2020learning}, OR-Tools-GLS      & \textbf{619.57}  & 4.14 & 619.21  & 8.30s & \textbf{1119.07}  & 6.67  & 1118.01 & 8.09s  \\
                      & random (1000) & 1646.83 & 6.13 & 1202.66 & -     & 8368.49  & 8.65  & 2897.99 & -      \\
                      & \citet{kool2019attention}, AM+TW(greedy) & 7615.69 & 2.00    & 1094.39 & 0.05s & 40245.40  & 2.00     & 2687.95 & 0.12s  \\
                      & \citet{kool2019attention}, AM+TW(sampl.) & 1572.31 & 6.56 & 1221.09 & 0.11s & 7712.35  & 9.34  & 2953.65 & 0.34s  \\
                      & \citet{kool2019attention}, AM+TW(t10240) & 1387.30  & 6.46 & 1170.49 & 0.90s & 6730.55  & 9.99  & 2954.76 & 2.70s  \\
                      & \citet{falkner2020learning}, JAMPR(greedy) & 674.72  & 4.32 & 626.80   & 0.11s & 1273.20   & 6.02  & 1126.74 & 0.25s  \\
                      & \citet{falkner2020learning}, JAMPR(sampl.) & \textbf{620.68}  & 4.19 & 602.33  & 0.92s & \textbf{1116.76}  & 5.64  & 1076.79 & 2.32s  \\ \midrule
\multirow{10}{*}{TW3} & \citet{falkner2020learning}, OR-Tools-AU       & 1317.81 & 4.07 & 637.15  & 1.07s & 2707.72  & 6.12  & 1121.57 & 20.63s \\
                      & \citet{falkner2020learning}, OR-Tools-GLS      & 1312.71 & 4.11 & 625.80   & 8.00s & 2753.66  & 6.18  & 1192.61 & 8.02s  \\
                      & random (1000) & 1409.35 & 3.34 & 953.48  & -     & 4407.58  & 6.92  & 2692.78 & -      \\
                      & \citet{kool2019attention}, AM+TW(greedy) & 3101.21 & 2.00    & 1094.39 & 0.05s & 26467.34 & 2.00     & 2687.95 & 0.12s  \\
                      & \citet{kool2019attention}, AM+TW(sampl.) & 1412.16 & 3.42 & 951.92  & 0.12s & 4161.24  & 7.15  & 2674.03 & 0.34s  \\
                      & \citet{kool2019attention}, AM+TW(t10240) & 1318.56 & 3.23 & 899.83  & 0.88s & 3941.47  & 6.77  & 2575.28 & 2.67s  \\
                      & \citet{falkner2020learning}, JAMPR(greedy) & 1002.81 & 1.00    & 733.01  & 0.10s & 3158.26  & 2.01  & 1347.72 & 0.23s  \\
                      &\citet{falkner2020learning}, JAMPR(sampl.) & \textbf{844.35}  & 1.39 & 660.48  & 0.78s & \textbf{1947.65}  & 2.29  & 1358.29 & 2.13s \\
\bottomrule
\end{tabular}
\begin{tablenotes}\footnotesize
\item[*] 
TW1 - problems with hard time windows, TW2 - problem with soft constraint for upper bound, TW3 - problem with soft constraints for both upper and lower bounds.
\end{tablenotes}
\end{table*}

\afterpage{
\begin{sidewaystable*}
\centering
\caption{The performance comparison on TSPLib.}
\label{tb:tsplib}
\resizebox{1\textwidth}{!}{%
\begin{tabular}{@{}ccccccccccc@{}}
\toprule
Instance                   & \begin{tabular}[c]{@{}l@{}}\citet{wu2021learning} \\(T=3k)\end{tabular} & \begin{tabular}[c]{@{}l@{}}\citet{ma2021learning}\\DACT\\ (T=3k)\end{tabular} & \begin{tabular}[c]{@{}l@{}}OR \\Tools \end{tabular}& \begin{tabular}[c]{@{}l@{}}\citet{kool2019attention}\\ (N=10k)\end{tabular} & \begin{tabular}[c]{@{}l@{}}\citet{kwon2020pomo} \\POMO \\$\times$8 augment\end{tabular} & \begin{tabular}[c]{@{}l@{}}\citet{wu2021learning} \\ (T=3k, M=1k)\end{tabular} & \begin{tabular}[c]{@{}l@{}}\citet{ma2021learning} \\ (T=10k)\end{tabular} & \begin{tabular}[c]{@{}l@{}}\citet{ma2021learning}\\$\times$4 augment\end{tabular} & \citet{da2021learning} & \begin{tabular}[c]{@{}l@{}}\citet{kim2021learning} \\AM + LCP\end{tabular} \\ \midrule
eil51  & 2.82\% & 1.64\% & 2.35\% & 2.11\% & 0.00\% & 1.17\% & \textbf{0.00\%} 
& \textbf{0.00\%} & 0.23\% & 0.73\%   \\

berlin52  & 6.34\% & \textbf{0.03\%} & 5.34\% & 1.67\% & \textbf{0.03\%} & 2.57\% & \textbf{0.03\%}    & \textbf{0.03\%} & 5.73\%   & 0.10\%   \\

st70 & 4.59\% & 0.44\% & 1.19\% & 2.22\% & \textbf{0.30\%} & 0.89\% & \textbf{0.30\%}         & \textbf{0.30\%} & 0.74\%   & 0.74\%   \\

eil76 & 6.88\% & 2.42\%  & 4.28\% & 3.35\% & \textbf{1.49\%} & 4.65\%  & 2.04\%     & 1.67\% & 2.60\%   & 1.64\%   \\

pr76 & 1.40\% & 1.02\% & 2.72\% & 2.84\% & 19.97\% & 1.37\% & \textbf{0.03\%}         & \textbf{0.03\%} & 2.60\%   & 0.44\%   \\

rat99                      & 17.18\%                   & 4.05\%                                                & 1.73\%   & 9.50\%                                                        & 7.51\%                                                      & 8.51\%                                                                    & 1.16\%                                                  & \textbf{0.74\%}                                                   & 14.62\%  & 6.67\%   \\
KroA100                    & 18.39\%                   & 0.86\%                                                & 0.78\%   & 79.49\%                                                       & 4.45\%                                                      & 2.08\%                                                                    & 0.63\%                                                  & \textbf{0.45\%}                                                   & 11.60\%  & 2.95\%   \\
KroB100                    & 19.97\%                   & 0.27\%                                                & 3.91\%   & 9.30\%                                                        & 5.83\%                                                      & 5.78\%                                                                    & \textbf{0.25\%}                                                  & \textbf{0.25\%}                                                   & 7.45\%   & 1.51\%   \\
KroC100                    & 22.14\%                   & 1.06\%                                                & 4.02\%   & 8.04\%                                                        & 6.55\%                                                      & 3.17\%                                                                    & \textbf{0.84\%}                                                  & \textbf{0.84\%}                                                   & 9.27\%   & 2.84\%   \\
KroD100                    & 16.33\%                   & 3.54\%                                                & 1.61\%   & 10.02\%                                                       & 8.74\%                                                      & 5.00\%                                                                    & 3.54\%                                                  & \textbf{0.12\%}                                                   & 9.58\%   & 1.97\%   \\
KroE100                    & 21.91\%                   & 2.17\%                                                & 2.40\%   & 3.10\%                                                        & 5.97\%                                                      & 3.29\%                                                                    & 1.95\%                                                  & \textbf{0.32\%}                                                   & 5.37\%   & 1.90\%   \\
rd100                      & 0.06\%                    & 0.08\%                                                & 3.53\%   & 1.93\%                                                        & 0.00\%                                                      & 0.06\%                                                                    & 0.06\%                                                  & \textbf{0.00\%}                                                   & 0.43\%   & 0.13\%   \\
eil101                     & 4.61\%                    & 3.66\%                                                & 5.56\%   & 3.97\%                                                        & \textbf{2.07\%}                                                     & 4.61\%                                                                    & 3.66\%                                                  & 2.86\%                                                    & 0.95\%   & 2.59\%   \\
lin105                     & 26.53\%                   & 3.41\%                                                & 3.09\%   & 32.13\%                                                       & 12.00\%                                                     & 2.48\%                                                                    & 3.35\%                                                  & \textbf{0.69\%}                                                   & 12.36\%  & 3.86\%   \\
pr107                      & 19.76\%                   & 5.86\%                                                & \textbf{1.74\%}  & 43.26\%                                                       & 5.66\%                                                      & 3.87\%                                                                    & 5.01\%                                                  & 3.81\%                                                    & -        & -        \\
pr124                      & 11.82\%                   & 1.56\%                                                & 5.91\%   & 4.41\%                                                        & \textbf{0.29\%}                                                     & 2.97\%                                                                    & 1.22\%                                                  & 1.22\%                                                    & 0.82\%   & 3.84\%   \\
bier127                    & 20.65\%                   & 4.08\%                                                & 3.76\%   & \textbf{1.71\%}                                                       & 60.56\%                                                     & 3.48\%                                                                    & 3.79\%                                                  & 2.46\%                                                    & 2.40\%   & 8.92\%   \\
ch130                      & 16.53\%                   & 6.63\%                                                & 2.85\%   & 2.96\%                                                        & \textbf{0.25\%}                                                     & 4.89\%                                                                    & 5.48\%                                                  & 1.93\%                                                    & 1.06\%   & 0.57\%   \\
pr136                      & 9.14\%                    & 5.54\%                                                & 5.62\%   & 4.90\%                                                        & \textbf{1.06\%}                                                     & 6.33\%                                                                    & 5.14\%                                                  & 4.54\%                                                    & 1.74\%   & 1.56\%   \\
pr144                      & 21.30\%                   & 3.44\%                                                & 1.28\%   & 8.77\%                                                        & \textbf{0.80\%}                                                     & 1.40\%                                                                    & 3.44\%                                                  & 2.49\%                                                    & 4.56\%   & 3.47\%   \\
ch150                      & 21.26\%                   & 3.60\%                                                & 3.08\%   & 3.45\%                                                        & \textbf{0.83\%}                                                     & 3.55\%                                                                    & 3.45\%                                                  & 1.23\%                                                    & -        & -        \\
KroA150                    & 17.80\%                   & 6.93\%                                                & 4.03\%   & 9.98\%                                                        & 13.15\%                                                     & 4.51\%                                                                    & 3.91\%                                                  & 3.91\%                                                   & 13.40\%  & \textbf{3.68\%}   \\
KroB150                    & 20.20\%                   & 6.10\%                                                & 5.52\%   & 9.87\%                                                        & 11.72\%                                                     & 5.40\%                                                                    & 4.10\%                                                  & \textbf{2.82\%}                                                   & 7.80\%   & 3.18\%   \\
pr152                      & 16.20\%                   & 4.48\%                                                & 2.92\%   & 13.47\%                                                       & 4.11\%                                                      & \textbf{2.17\%}                                                                   & 3.59\%                                                  & 3.59\%                                                    & 2.20\%   & 2.52\%   \\
u159                       & 21.97\%                   & 6.84\%                                                & 8.79\%   & 7.38\%                                                        & 2.19\%                                                     & 7.67\%                                                                    & 5.86\%                                                  & 3.16\%                                                   & \textbf{1.51\%}   & 10.84\%  \\
rat195                     & 25.40\%                   & 6.93\%                                                & \textbf{2.84\%}  & 16.57\%                                                       & 29.06\%                                                     & 9.90\%                                                                    & 5.81\%                                                  & 4.99\%                                                    & 27.21\%  & 10.81\%  \\
d198                       & 13.83\%                   & 12.27\%                                               & \textbf{1.16\%}  & 331.58\%                                                      & 45.98\%                                                     & 4.99\%                                                                    & 10.74\%                                                 & 8.75\%                                                    & -        & -        \\
KroA200                    & 22.44\%                   & 3.60\%                                                & 1.27\%   & 15.64\%                                                       & 20.00\%                                                     & 7.01\%                                                                    & 1.52\%                                                  & \textbf{1.25\%}                                                   & 10.74\%  & 6.14\%   \\
KroB200                    & 23.69\%                   & 10.51\%                                               & \textbf{3.67\%}  & 18.54\%                                                       & 21.06\%                                                     & 7.05\%                                                                    & 6.28\%                                                  & 5.66\%                                                    & -        & -        \\ \midrule
Avg. Gap, {[}50,100)    & 6.53\%                    & 1.60\%                                                & 2.93\%   & 3.61\%                                                        & 4.88\%                                                      & 3.19\%                                                                    & 0.59\%                                                  & \textbf{0.46\%}                                                   & 4.42\%   & 1.72\%   \\
Avg. Gap, {[}100,150)   & 9.69\%                    & 3.01\%                                                & 3.29\%   & 15.29\%                                                       & 8.16\%                                                      & 3.53\%                                                                    & 2.74\%                                                  & \textbf{1.57\%}                                                   & 5.20\%   & 2.78\%   \\
Avg. Gap, {[}150,200{]} & 12.76\%                   & 6.81\%                                                & 3.70\%   & 47.39\%                                                       & 16.45\%                                                     & 5.81\%                                                                    & 5.03\%                                                  & \textbf{3.93\%}                                                   & 10.48\%  & 6.20\%   \\ \midrule
Avg. Gap all  & 15.56\%                   & 3.90\%                                                & 3.34\%   & 22.83\%                                                       & 10.06\%                                                     & 4.17\%                                                                    & 3.01\%                                                  & \textbf{2.07\%}                                                   & 6.28\%   & 3.34\%   \\ \bottomrule
\end{tabular}}
\end{sidewaystable*}
}

\afterpage{
\begin{sidewaystable*}
\centering
\caption{The performance comparison  on CVRPLib.}
\label{tb:cvrplib}
\resizebox{0.95\textwidth}{!}{%
\begin{tabular}{@{}ccccccccccc@{}}
\toprule
Instance      & \begin{tabular}[c]{@{}l@{}}Depot \\ Type \end{tabular}   & \begin{tabular}[c]{@{}l@{}}Customer \\ Type \end{tabular}  & \begin{tabular}[c]{@{}l@{}}\citet{wu2021learning} \\(T=5k)\end{tabular}  & \begin{tabular}[c]{@{}l@{}}\citet{ma2021learning} \\ (T=5k)\end{tabular} & \begin{tabular}[c]{@{}l@{}}OR\\ Tools \end{tabular} & \begin{tabular}[c]{@{}l@{}}\citet{kool2019attention}\\ (N=10k)\end{tabular} & \begin{tabular}[c]{@{}l@{}}\citet{kwon2020pomo} \\$\times$8 augment\end{tabular} & \begin{tabular}[c]{@{}l@{}}\citet{wu2021learning} \\ (T=5k, M=100)\end{tabular} & \begin{tabular}[c]{@{}l@{}}\citet{ma2021learning} \\ (T=10k)\end{tabular} & \begin{tabular}[c]{@{}l@{}}\citet{ma2021learning} \\$\times$6 augment\end{tabular} \\
\midrule
X-n101-k25    & R             & R              & 7.70\%                    & 2.09\%                                                 & 6.57\%   & 32.95\%                                                       & 3.64\%                                                      & 5.60\%                                                                      & 1.86\%                                                  & \textbf{1.47\%}                                                    \\
X-n106-k14    & E             & C              & 4.86\%                    & 2.93\%                                                 & 3.72\%   & 6.78\%                                                        & \textbf{1.85\%}                                                     & 2.83\%                                                                      & 2.75\%                                                  & 1.87\%                                                     \\
X-n110-k13    & C             & R              & 6.39\%                    & 1.43\%                                                 & 7.87\%   & 3.15\%                                                        & 2.05\%                                                      & 4.40\%                                                                      & 0.87\%                                                  & \textbf{0.13\%}                                                    \\
X-n115-k10    & C             & R              & 13.32\%                   & 3.29\%                                                 & 4.50\%   & 7.52\%                                                        & 3.49\%                                                      & 5.19\%                                                                      & 3.26\%                                                  & \textbf{1.68\%}                                                    \\
X-n120-k6     & E             & RC             & 16.16\%                   & 3.50\%                                                 & 6.83\%   & 4.54\%                                                        & \textbf{2.12\%}                                                     & 5.56\%                                                                      & 3.20\%                                                  & 2.38\%                                                     \\
X-n125-k30    & R             & C              & 8.79\%                    & 6.51\%                                                 & 5.63\%   & 35.16\%                                                       & 7.14\%                                                      & \textbf{4.71\%}                                                                     & 5.47\%                                                  & 5.44\%                                                     \\
X-n129-k18    & E             & RC             & 11.01\%                   & 2.93\%                                                 & 8.37\%   & 4.00\%                                                        & \textbf{0.97\%}                                                     & 4.63\%                                                                      & 2.55\%                                                  & 2.55\%                                                     \\
X-n134-k13    & R             & C              & 16.06\%                   & 6.98\%                                                 & 21.61\%  & 20.13\%                                                       & 4.22\%                                                      & 8.88\%                                                                      & 5.56\%                                                  & \textbf{2.63\%}                                                    \\
X-n139-k10    & C             & R              & 14.99\%                   & 2.54\%                                                 & 12.02\%  & 4.30\%                                                        & 2.28\%                                                      & 4.90\%                                                                      & 2.16\%                                                  & \textbf{2.08\%}                                                    \\
X-n143-k7     & E             & R              & 20.20\%                   & 7.80\%                                                 & 11.27\%  & 8.88\%                                                        & \textbf{2.79\%}                                                     & 6.61\%                                                                      & 6.47\%                                                  & 3.55\%                                                     \\
X-n148-k46    & R             & RC             & 16.38\%                   & 2.69\%                                                 & 7.80\%   & 79.53\%                                                       & 19.88\%                                                     & 3.60\%                                                                      & \textbf{2.22\%}                                                  & \textbf{2.22\%}                                                    \\
X-n153-k22    & C             & C              & 22.94\%                   & 11.06\%                                                & 8.01\%   & 78.11\%                                                       & 12.16\%                                                     & \textbf{4.53\%}                                                                     & 9.02\%                                                  & 6.53\%                                                     \\
X-n157-k13    & R             & C              & 17.15\%                   & 4.64\%                                                 & \textbf{2.57\%}  & 16.30\%                                                       & 2.79\%                                                      & 3.60\%                                                                      & 4.44\%                                                  & 3.12\%                                                     \\
X-n162-k11    & C             & RC             & 19.16\%                   & 4.43\%                                                 & 6.31\%   & 6.37\%                                                        & 4.77\%                                                      & 5.26\%                                                                      & 3.04\%                                                  & \textbf{2.62\%}                                                    \\
X-n167-k10    & E             & R              & 18.52\%                   & 5.37\%                                                 & 9.34\%   & 8.41\%                                                        & 4.05\%                                                      & 8.27\%                                                                      & 4.28\%                                                  & \textbf{3.47\%}                                                    \\
X-n172-k51    & C             & RC             & 12.06\%                   & 6.23\%                                                 & 10.74\%  & 85.37\%                                                       & 21.99\%                                                     & 4.36\%                                                                      & 5.27\%                                                  & \textbf{3.41\%}                                                    \\
X-n176-k26    & E             & R              & 19.49\%                   & 10.29\%                                                & 8.99\%   & 20.39\%                                                       & 10.27\%                                                     & 6.16\%                                                                      & 8.07\%                                                  & \textbf{5.93\%}                                                    \\
X-n181-k23    & R             & C              & 6.27\%                    & 3.41\%                                                 & 2.94\%   & 6.45\%                                                        & \textbf{2.08\%}                                                      & \textbf{2.08\%}                                                                      & 2.42\%                                                  & \textbf{2.08\%}                                                    \\
X-n186-k15    & R             & R              & 17.71\%                   & 5.99\%                                                 & 7.75\%   & 6.01\%                                                        & \textbf{2.15\%}                                                     & 7.65\%                                                                      & 5.30\%                                                  & 4.94\%                                                     \\
X-n190-k8     & E             & C              & 18.64\%                   & 7.97\%                                                 & \textbf{6.53\%}  & 46.61\%                                                       & 9.25\%                                                      & 6.78\%                                                                      & 6.73\%                                                  & 6.73\%                                                     \\
X-n195-k51    & C             & RC             & 17.04\%                   & 7.00\%                                                 & 13.76\%  & 79.26\%                                                       & 9.23\%                                                      & 4.47\%                                                                      & 4.54\%                                                  & \textbf{4.36\%}                                                    \\
X-n200-k36    & R             & C              & 9.60\%                    & 5.93\%                                                 & \textbf{4.15\%}  & 26.25\%                                                       & 5.01\%                                                      & 4.26\%                                                                      & 5.87\%                                                  & 5.86\%                                                     \\
X-n129-k18    & E             & RC             & 11.01\%                   & 2.93\%                                                 & 8.37\%   & 4.00\%                                                        & \textbf{0.97\%}                                                     & 4.63\%                                                                      & 2.55\%                                                  & 2.55\%                                                     \\
X-n134-k13    & R             & C              & 16.06\%                   & 6.98\%                                                 & 21.61\%  & 20.13\%                                                       & 4.22\%                                                      & 8.88\%                                                                      & 5.56\%                                                  & \textbf{2.63\%}                                                    \\
X-n139-k10    & C             & R              & 14.99\%                   & 2.54\%                                                 & 12.02\%  & 4.30\%                                                        & 2.28\%                                                      & 4.90\%                                                                      & 2.16\%                                                  & \textbf{2.08\%}                                                    \\
X-n143-k7     & E             & R              & 20.20\%                   & 7.80\%                                                 & 11.27\%  & 8.88\%                                                        & \textbf{2.79\%}                                                     & 6.61\%                                                                      & 6.47\%                                                  & 3.55\%                                                     \\
X-n148-k46    & R             & RC             & 16.38\%                   & 2.69\%                                                 & 7.80\%   & 79.53\%                                                       & 19.88\%                                                     & 3.60\%                                                                      & \textbf{2.22\%}                                                  & \textbf{2.22\%}                                                    \\
X-n153-k22    & C             & C              & 22.94\%                   & 11.06\%                                                & 8.01\%   & 78.11\%                                                       & 12.16\%                                                     & \textbf{4.53\%}                                                                     & 9.02\%                                                  & 6.53\%                                                     \\
X-n157-k13    & R             & C              & 17.15\%                   & 4.64\%                                                 & \textbf{2.57\%}  & 16.30\%                                                       & 2.79\%                                                      & 3.60\%                                                                      & 4.44\%                                                  & 3.12\%                                                     \\
X-n162-k11    & C             & RC             & 19.16\%                   & 4.43\%                                                 & 6.31\%   & 6.37\%                                                        & 4.77\%                                                      & 5.26\%                                                                      & 3.04\%                                                  & \textbf{2.62\%}                                                    \\
X-n167-k10    & E             & R              & 18.52\%                   & 5.37\%                                                 & 9.34\%   & 8.41\%                                                        & 4.05\%                                                      & 8.27\%                                                                      & 4.28\%                                                  & \textbf{3.47\%}                                                    \\
X-n172-k51    & C             & RC             & 12.06\%                   & 6.23\%                                                 & 10.74\%  & 85.37\%                                                       & 21.99\%                                                     & 4.36\%                                                                      & 5.27\%                                                  & \textbf{3.41\%}                                                    \\
X-n176-k26    & E             & R              & 19.49\%                   & 10.29\%                                                & 8.99\%   & 20.39\%                                                       & 10.27\%                                                     & 6.16\%                                                                      & 8.07\%                                                  & \textbf{5.93\%}                                                    \\
X-n181-k23    & R             & C              & 6.27\%                    & 3.41\%                                                 & 2.94\%   & 6.45\%                                                        & 2.08\%                                                      & 2.08\%                                                                      & 2.42\%                                                  & \textbf{2.08\%}                                                    \\
X-n186-k15    & R             & R              & 17.71\%                   & 5.99\%                                                 & 7.75\%   & 6.01\%                                                        & \textbf{2.15\%}                                                     & 7.65\%                                                                      & 5.30\%                                                  & 4.94\%                                                     \\
X-n190-k8     & E             & C              & 18.64\%                   & 7.97\%                                                 & \textbf{6.53\%}  & 46.61\%                                                       & 9.25\%                                                      & 6.78\%                                                                      & 6.73\%                                                  & 6.73\%                                                     \\
X-n195-k51    & C             & RC             & 17.04\%                   & 7.00\%                                                 & 13.76\%  & 79.26\%                                                       & 9.23\%                                                      & 4.47\%                                                                      & 4.54\%                                                  & \textbf{4.36\%}                                                    \\
X-n200-k36    & R             & C              & 9.60\%                    & 5.93\%                                                 & \textbf{4.15\%}  & 26.25\%                                                       & 5.01\%                                                      & 4.26\%                                                                      & 5.87\%                                                  & 5.86\%                                                     \\\midrule
\multicolumn{3}{l}{Avg. Gap for {[}100,150)}   & 12.35\%                   & 3.88\%                                                 & 8.74\%   & 18.81\%                                                       & 4.58\%                                                      & 5.17\%                                                                      & 3.31\%                                                  & \textbf{2.36\%}                                                    \\
\multicolumn{3}{l}{Avg. Gap for {[}150,200{]}} & 16.24\%                   & 6.57\%                                                 & 7.37\%   & 34.50\%                                                       & 7.61\%                                                      & 5.22\%                                                                      & 5.36\%                                                  & \textbf{4.46\%}                                                    \\ \midrule
\multicolumn{3}{l}{Avg. Gap for all} & 14.29\%                   & 5.23\%                                                 & 8.06\%   & 26.66\%                                                       & 6.10\%                                                      & 5.20\%                                                                      & 4.33\%                                                  & \textbf{3.41\%}                                                    \\
              &               &                &                           &                                                        &          &                                                               &                                                             &                                                                             &                                                         &     \\ \bottomrule                               

\end{tabular}}
\end{sidewaystable*}
}

\subsection{Multi-vehicle Routing}\label{sec:rps}
Multiple vehicle routing problems can be also modeled using the above single-agent reinforcement learning formulation. For instance, in routing $M$ homogeneous vehicles with the objective to minimize the total distance traveled, the trained reinforcement learning agent solution can be split into $M$ vehicles, where each vehicle is allowed to return to the depot only once. 
However, in VRPs, all the vehicles share a common goal of serving customers with the least total costs. Therefore, the idea of using \emph{a centralized controller}, which is responsible for routing all vehicles in coordination in order to achieve a common goal has been used in many studies. 

In \emph{a centralized controller} setting, the controller is the agent, which takes actions for all vehicles at each time step. Formally, the central controller has action state $\mathbb{A}_{t+1} = \{A^1_{t+1}, \dots, A^M_{t+1}\}$, where $M$ is the number of vehicles and $A^m_{t+1} \in \mathbb{V} $ for all $m= 1, \dots, M$. The current state $S_t = \{\mathbb{A}_t, \vec{l_t}, Y_t\}$ consists from tuple indicating the last selected actions of all vehicles, the vector of remaining loads for each vehicle and the set of customers served by any vehicle. Then the central controller is trained to efficiently construct routes for all vehicles denoted as $p = [p_0, \dots, p_T]$, where $p_t = [a^1_t, \dots, a^M_t]$. Then for each vehicle $m$ we have a route $p^m = [a^m_0, \dots a^m_T]$. In case when the objective is to minimize the total time spent to serve all customers, the central controller aims to minimize the total reward defined as:
\begin{align}
    R = \min_{p}\{\text{max} \{c^1(p^1), \dots, c^m(p^m)\}\}
\end{align}
where $c^m(p^m) = \sum^{T-1}_{t=0} \Delta t^m_{a^m_t, a^m_{t+1}}$, where $\Delta t^m_{a^m_t, a^m_{t+1}}$ is travelling time of vehicle $m$ from node $a^m_t$ to node $a^m_{t+1}$. 
Unlike the MARL formulation, the central controller observes the entire environment and has full knowledge about the state of each vehicle, which allows it to promote coordinated routing for the common goal of both homogeneous fleets and heterogeneous vehicles. For instance, all vehicles have common action space for multiple vehicles operating in a shared graph. Thus, without coordination, several vehicles may visit the same customer. The central controller, which assigns actions for each vehicle prevents such inefficiencies. For instance,  \cite{vera2019deep, bogyrbayeva2021reinforcement, van2016deep, qin2021novel} and \cite{gutierrez2019selecting, zhang2020deep, lin2021deep} use the centralized controller to solve mCVRP and mVRPWT with homogeneous set of vehicles. \cite{james2019online} and \cite{Lin2019EfficientCM} propose deep reinforcement learning to route multiple vehicles with uncertain demand. Another emerging topic is to route a set of heterogeneous vehicles such as trucks and drones \cite{yoon2021deep, chen2019learning} that also rely on the centralized controller. 

On the other hand, MARL aims to train fully independent agents which cooperate together for a common goal. Formally each agent $m$ has its local observation  denoted as $o^m_t$, which only includes partial information about the environment. Then each agent is trained to select action $A^m_{t+1}$ given its observation $o^m_t$ and its state $s^m_t$. In order to promote coordinated routing, communication messages are allowed between agents.  Then each agent is trained to construct its route $p^m=[a^m_0, \dots, a^m_T]$, that contributes to the common reward:
\begin{align}
    R = \sum_{m=1}^M c^m(p^m)
\end{align}
where $c^m(p^m)$ is a cost function. The general framework to apply MARL  for routing multiple vehicles is presented in  \cite{silva2019reinforcement}. The MARL formulations are suitable in dynamic environments when the customer demand is observed over time. For instance, in the case of online routing, vehicles trained in a MARL setting have the ability to make independent decisions to serve customers.  For instance, \cite{sykora2020multi} presents the MARL formulation to solve Multi-agent Mapping Problem to satisfy demand in known locations with unknown quantities.
Table \ref{tb:multi} summarizes studies focused on routing multiple vehicles. 

\begin{table*}[]
\caption{The summary of studies to solve multi-vehichle routing problems. DSCVRPTW - Dynamic and Stochastic CVRPTW, SCVRPTW-Stochastic CVRPTW, MMHCVRP-min-max heterogeneous CVRP, MSHCVR-min-sum heterogeneous CVRP. }
\centering 
\label{tb:multi}
\begin{tabular}{>{\centering}p{0.22\linewidth}  >{\centering}p{0.2\linewidth}  >{\centering}p{0.065\linewidth}  >{\centering}p{0.065\linewidth} >{\centering}p{0.065\linewidth} >{\centering}p{0.065\linewidth} >{\centering}p{0.065\linewidth} >{\centering\arraybackslash}p{0.065\linewidth}
}
\toprule
Name                                                                                                        & Problems                                                                                               & Multi-\\agent & Central \\Controller &  Hetero-\\geneous  &  Homo-\\geneous  & Offline & Online \\ \midrule
                                                                                                           
\citet{james2019online}              & Online DVRP                                                                &             & \checkmark           & \checkmark               &            &         & \checkmark         \\
\citet{vera2019deep}                                 & mCVRP                                                                                                  &             & \checkmark           & \checkmark            &            & \checkmark       &        \\
\citet{zhang2020multi}      & mVRP with soft TW                                          &             & \checkmark           &              & \checkmark          & \checkmark       &        \\
\citet{sykora2020multi}                                                              &  Multi-agent Mapping Problem                        & \checkmark           &             &              & \checkmark          & \checkmark       &        \\
\citet{Lin2019EfficientCM}         &  online ride-sharing &             & \checkmark           &              & \checkmark          &         & \checkmark      \\
\citet{falkner2020learning}                      & mCVRPTW                                                                              &             & \checkmark           &              & \checkmark          & \checkmark       &        \\
\citet{qin2021novel}           & VRP                                                                                      &             & \checkmark           & \checkmark            &            & \checkmark       &        \\
\citet{silva2019reinforcement} & mVRPTW                                                                                        & \checkmark           &             &              & \checkmark          & \checkmark       &        \\
\citet{van2020complex}                                            & mCVRP                                                                                            &             & \checkmark           &              & \checkmark          & \checkmark       &   \\

\citet{bogyrbayeva2021reinforcement} & mCVRP with charging & & \checkmark & &\checkmark & \checkmark & \\

\citet{yoon2021deep} & TSP with Drone & &\checkmark &\checkmark & &\checkmark & 
\\

\citet{chen2022deep} & SDDPVD & &\checkmark &\checkmark & & &\checkmark 
\\

\citet{lin2021deep} & mEVRPTW & &\checkmark & &\checkmark &\checkmark & 
\\ 

\citet{gutierrez2019selecting} & mVRPTW & &\checkmark & &\checkmark &\checkmark & 
\\

\citet{bono2020solving} & \begin{tabular}[c]{@{}c@{}}mDSCVRPTW, mSCVRPTW\\  \end{tabular} & & \checkmark 
& &\checkmark &  & \checkmark \\

\citet{licvrp2021deep} & MMHCVRP, MSHCVRP & & \checkmark 
&\checkmark & &  & \checkmark \\
\bottomrule
\end{tabular}
\end{table*}

\section{The  Future Research Directions}\label{sec:future}

This section discusses future research directions based on the presented survey. 

\paragraph{Generalization}
The generalization of learning methods to problems of different graph sizes is one of the primary and common challenges faced by many proposed models. For instance, to produce competitive solutions, separate sets of models need to be trained from scratch for each graph size \cite{kool2019attention, nazari2018reinforcement, bello2016neural, dai2017learning}. This leads to a substantial total training time, which can be sped up by advanced hardware equipped with multiple GPUs. Nevertheless, developing training techniques to transfer learning from small-sized graphs to large-sized graphs has not been studied extensively except in a few studies. \cite{fu2021generalize} proposes to learn heatmaps for TSP on small sizes and apply them to arbitrarily large instances. Also, \cite{joshi2021learning} investigates the generalization capabilities of supervised and reinforcement learning methods in end-to-end settings on TSP. The paper has shown that reinforcement learning is better than supervised learning because it does not learn from an existing solution. The study demonstrates that models trained on various graph sizes generalize better than the models trained on solo-sized graphs. 

Scaling to large instances is an especially severe issue for end-to-end methods because training such models requires a large time budget and efficient hardware. On the contrary, hybrid models have demonstrated the potential to handle large-scale routing problems. For instance, \cite{xing2020graph} presents a model to solve TSP with 1,000 nodes. \cite{li2021learning} speeds up the current solvers to solve VRPs with 500 and up to 3,000 nodes. \cite{gao2020learn}' experiments indicate the competitiveness of their hybrid method with the construction heuristics for VRPWT with 400 nodes. 

Even though presenting the performance of the proposed models on real-world datasets has become more common in the recent years \cite{wu2021learning, ma2019combinatorial, kim2021learning, bogyrbayeva2021reinforcement, yoon2021deep}, the generalization of the learning-based models to real-world data is not well studied yet. Recently there have been some attempts \cite{accorsi2022guidelines} to establish well-accepted benchmark libraries and guidelines to compare the computational results of machine learning models to solve combinatorial optimization problems with the operations research methods. However, the fair comparison of the learning-based models among themselves and against non-learning-based methods still remains a complex and challenging task. 

\paragraph{Improving Models and Solution Methods}
Complex models equipped with advanced deep neural techniques and numerous parameters to train on expensive hardware have been proposed to solve routing problems. However, the solutions produced by learning methods in the best cases are comparable with the existing non-learning methods as reported in Tables \ref{tb:tsp}, \ref{tb:cvrp}, \ref{tb:cvrplarge} and \ref{tb:vrptw}. Learning methods require the development of simple yet powerful models that can compete with traditional methods.

Currently, there are strong assumptions about the inputs to the models. For instance, both end-to-end and hybrid models work with complete graphs with known coordinates of the nodes and distances.   However, practically important VRPs focus on the characteristics of the road networks such as costs of edges \cite{corberan2021arc} without complete information about a graph. This indicates a strong need to develop models that can generalize to edge features only. 

In addition to inputs, the central question of why the presented models work remains largely unanswered. The studies present the superiority of their proposed models based on the computational studies conducted mostly on random datasets. However, the discussions over their choice of deep learning models that fit the properties of VRPs are limited. The comparison studies investigating properties of deep learning models for specific tasks such as graph embedding and a solution decoding for VRPs are needed. 

\paragraph{Incorporating Uncertainty and Online Routing}
The power of learning models lies in their ability to generalize over data distribution. This advantage can be well exploited to incorporate dynamic elements in routing problems, which is not easy to do with the traditional methods \cite{ulmer2019offline}. For instance, online routing has become more critical with the increased demand for on-demand services such as parcel delivery \cite{james2019online, chen2022deep}, ride-sharing \cite{Lin2019EfficientCM, sykora2020multi} and fractional ownership \cite{bogyrbayeva2021iterative, takalloo2021solving}. 
The development of green logistic systems, including electrical, autonomous vehicles, and drones with uncertain charging times, can also be incorporated into the development of learning-based routing problems.  

However, routing multiple vehicles even with static inputs is challenging \cite{bogyrbayeva2021reinforcement}.  
The majority of the surveyed studies focus on routing either a single vehicle or designed to solve the problems with the additive objectives such as distance. The fundamental problem of routing multiple vehicles that share a common action space requires developing novel or proper adjustments to existing algorithms. 

In summary, learning-based methods, either in their pure or hybrid forms, possess advantages such as solution speed and incorporating uncertainty in overcoming challenges faced by the urban transportation systems through solving vehicle routing problems.

\bibliographystyle{IEEEtranN-ital}
\bibliography{sample}

\clearpage

\end{document}